\documentclass{article}

\usepackage[preprint]{neurips_2022}

\usepackage[utf8]{inputenc} 
\usepackage[T1]{fontenc}    
\usepackage{hyperref}       
\usepackage{url}            
\usepackage{booktabs}       
\usepackage{amsfonts}       
\usepackage{nicefrac}       
\usepackage{microtype}      
\usepackage{xcolor}         
\usepackage{bm}
\usepackage{amsmath}
\usepackage{graphicx}
\usepackage{comment}
\usepackage{listings}
\newcommand\bmt[1]{\bm{\tilde{#1}}}

\title{Multi-Sample Training for Neural Image Compression}

\author{%
Tongda Xu\textsuperscript{\rm 1,2}, Yan Wang\textsuperscript{\rm 1,2} \thanks{Yan Wang is the corresponding author.}, Dailan He\textsuperscript{\rm 1}, Chenjian Gao\textsuperscript{\rm 1,3}, Han Gao\textsuperscript{\rm 1,4}, \\\textbf{Kunzan Liu}\textsuperscript{\rm 1,2}, \textbf{Hongwei Qin}\textsuperscript{\rm 1} \\
  \textsuperscript{\rm 1}SenseTime Research, \textsuperscript{\rm 2}Tsinghua University,\\\textsuperscript{\rm 3}Beihang University, \textsuperscript{\rm 4}University of Electronic Science and Technology of China\\
  \texttt{\{xutongda, wangyan\}@air.tsinghua.edu.cn,} \\
  \texttt{\{hedailan, gaochenjian, gaohan1, liukunzan, qinhongwei\}@sensetime.com}
}

\begin{document}

\maketitle

\begin{abstract}
    This paper considers the problem of lossy neural image compression (NIC). Current state-of-the-art (sota) methods adopt uniform posterior to approximate quantization noise, and single-sample pathwise estimator to approximate the gradient of evidence lower bound (ELBO). In this paper, we propose to train NIC with multiple-sample importance weighted autoencoder (IWAE) target, which is tighter than ELBO and converges to log likelihood as sample size increases. First, we identify that the uniform posterior of NIC has special properties, which affect the variance and bias of pathwise and score function estimators of the IWAE target. Moreover, we provide insights on a commonly adopted trick in NIC from gradient variance perspective. Based on those analysis, we further propose multiple-sample NIC (MS-NIC), an enhanced IWAE target for NIC. Experimental results demonstrate that it improves sota NIC methods. Our MS-NIC is plug-and-play, and can be easily extended to other neural compression tasks.
\end{abstract}

\section{Introduction}

Latent variable-based lossy neural image compression (NIC) has witnessed significant success. The majority of NIC follows the framework proposed by \citet{balle2017end}: For encoding, the original image $\bm{x}$ is transformed into $\bm{y}$ by the encoder. Then $\bm{y}$ is scalar-quantized into integer $\bm{\bar{y}}$, estimated with an entropy model $p(\bm{\bar{y}})$ and coded. For decoding, $\bm{\bar{y}}$ is transformed back by the decoder to obtain reconstructed $\bm{\bar{x}}$. The optimization target of NIC is R-D cost: $R + \lambda D$. $R$ denotes the bitrate of $\bm{\bar{y}}$, $D$ denotes the distortion between $\bm{x}$ and $\bm{\bar{x}}$, and $\lambda$ denotes the hyper-parameter controlling their trade-off. During training, the quantization $\bm{\bar{y}} = \lfloor\bm{y}\rceil$ is relaxed with $\bm{\tilde{y}} = \bm{y} + \bm{\epsilon}$ to simulate the quantization noise. And $\bm{\epsilon}$ is fully factorized uniform noise $\bm{{\epsilon}} \sim p(\bm{\epsilon}) = \prod \mathcal{U}(-\frac{1}{2}, +\frac{1}{2})$.

\citet{balle2017end} further recognises that such training framework is closely related to variational inference. Indeed, the above process can be formulated as a graphic model $\bm{x} \leftarrow \bmt{y}$. During encoding, $\bm{x}$ is transformed into variational parameter $\bm{y}$ by inference model (encoder), and $\bm{\tilde{y}}$ is sampled from variational posterior $q(\bm{\tilde{y}}|\bm{x})$, which is a unit unifrom distribution centered in $\bm{y}$. The prior likelihood $p(\bmt{y})$ is computed, and $\bmt{y}$ is transformed back by the generative model (decoder) to compute the likelihood $p(\bm{x}|\bm{\tilde{y}})$. Under such formulation, the prior is connected to the bitrate, the likelihood is connected to the distortion, and the posterior likelihood is connected to the bits-back bitrate (See Appendix.~\ref{sec:bb}), which is $0$ in NIC. Finally, the evidence lower bound (ELBO) is the negative $R+\lambda D$ target (Eq.~\ref{eq:gg17elbo}). Denote the transform function $\bmt{y}(\bm{\epsilon}; \phi) = \bm{y} + \bm{\epsilon}$, and sampling $\bmt{y} \sim q(\bmt{y}|\bm{x})$ is equivalent to transforming $\bm{\epsilon}$ through $\bmt{y}(\bm{\epsilon}; \phi)$. Then the gradient of ELBO is estimated via pathwise estimator with single-sample Monte Carlo (Eq.~\ref{eq:gg17elbograd}). This is the same as SGVB-1 \citep{kingma2013auto}.

\begin{equation}
  \begin{array}{l}
  \mathcal{L} = -(R+\lambda D)= \mathbb{E}_{q(\bm{\tilde{y}}|\bm{x})}[\underbrace{\log p(\bm{x}|\bm{\tilde{y}})}_{\textrm{\scriptsize{- distortion}}} + \underbrace{\log p(\bm{\tilde{y}})}_{\textrm{\scriptsize{- rate}}} \underbrace{-\log q(\bm{\tilde{y}}|\bm{x})}_{\textrm{\scriptsize{bits-back rate: 0}}}]
  \end{array}
\label{eq:gg17elbo}
\end{equation}

\begin{equation}
  \begin{array}{l}
  \nabla_{\phi}\mathcal{L} = \mathbb{E}_{p(\bm{\epsilon})}[\nabla_{\phi}(\log \frac{p(\bm{x}, \bmt{y}(\bm{\epsilon}; \phi))}{q(\bmt{y}(\bm{\epsilon}; \phi)|\bm{x})})] 
  \approx \nabla_{\phi}\log \frac{p(\bm{x}, \bmt{y}(\bm{\epsilon}; \phi))}{q(\bmt{y}(\bm{\epsilon}; \phi)|\bm{x})} 
  \end{array}
\label{eq:gg17elbograd}
\end{equation}

\citet{balle2018variational} further extends this framework into a two-level hierarchical structure, with graphic model $\bm{x} \leftarrow \bmt{y} \leftarrow \bmt{z}$. The variational posterior is fully factorized uniform distribution $\mathcal{U}(\bm{y}-\frac{1}{2}, \bm{y}+\frac{1}{2})\mathcal{U}(\bm{z}-\frac{1}{2}, \bm{z}+\frac{1}{2})$ To simulate the quantization noise. And $\bm{y}, \bm{z}$ denote outputs of their inference networks.

\begin{equation}
  \begin{array}{l}
  \mathcal{L} = \mathbb{E}_{q(\bm{\tilde{y}}, \bmt{z}|\bm{x})}[\underbrace{\log p(\bm{x}|\bm{\tilde{y}})}_{\textrm{\scriptsize{- distortion}}} + \underbrace{\log p(\bm{\tilde{y}}|\bm{\tilde{z}}) + \log p(\bm{\tilde{z}})}_{\textrm{\scriptsize{- rate}}} \underbrace{-\log q(\bm{\tilde{y}}|\bm{x}) -\log q(\bm{\tilde{z}}|\bm{\tilde{y}})}_{\textrm{\scriptsize{bits-back rate: 0}}}]
  \end{array}
\label{eq:gg18elbo}
\end{equation}

The majority of later NIC follows this hierarchical latent framework \citep{minnen2018joint, cheng2020learned}. Some focus on more expressive network architectures \citep{zhu2021transformer, xie2021enhanced}, some stress better context models \citep{minnen2020channel, he2021checkerboard, guo2021causal}, and some emphasize semi-amortization inference \citep{yang2020improving}. However, there is little research on multiple-sample methods, or other techniques for a tighter ELBO. 

On the other hand, IWAE \citep{burda2016importance} has been successful in density estimation. Specifically, IWAE considers a multiple-sample lowerbound $\mathcal{L}_k$ (Eq.~\ref{eq:iwaeelbo}), which is tighter than its single-sample counterpart. The benefit of such bound is that the implicit distribution defined by IWAE approaches true posterior as $k$ increases \citep{cremer2017reinterpreting}. This suggests that its variational posterior is less likely to collapse to a single mode of true posterior, and the learned representation is richer. The gradient of $\mathcal{L}_k$ is computed via pathwise estimator. Denote the exponential ELBO sample as $w_i$, its reparameterization as $w(\bm{\epsilon}_i; \phi)$, and its weight $\tilde{w}_{i} = \frac{w_{i}}{\sum w_{j}}$. Then $\nabla_{\phi}\mathcal{L}_k$ has the form of importance weighted sum (Eq.~\ref{eq:iwaeelbograd}).

\begin{equation}
  \begin{array}{l}
  \mathcal{L}_k = \mathbb{E}_{q(\bmt{y}_{1:k}|\bm{x})}[\log \frac{1}{k}\overset{k}{\underset{i}{\sum}} \underbrace{\frac{p(\bm{x}, \bmt{y}_i)}{q(\bmt{y}_{i}|\bm{x})}}_{w_i}] = \mathbb{E}_{p(\bm{\epsilon}_{1:k})}[\log \frac{1}{k}\overset{k}{\underset{i}{\sum}} \underbrace{\frac{p(\bm{x}, \bmt{y}(\bm{\epsilon}_{i};\phi))}{q(\bmt{y}(\bm{\epsilon}_{i}; \phi)|\bm{x})}}_{w(\bm{\epsilon}_i; \phi)}]
  \end{array}
\label{eq:iwaeelbo}
\end{equation}

\begin{equation}
  \begin{array}{l}
  \nabla_{\phi}\mathcal{L}_k = \mathbb{E}_{p(\bm{\epsilon}_{1:k})}[\overset{k}{\underset{i}{\sum}} \tilde{w}_{i} \nabla_{\phi} \log w(\bm{\epsilon}_{i}; \phi)] \approx \overset{k}{\underset{i}{\sum}} \tilde{w}_{i} \nabla_{\phi} \log w(\bm{\epsilon}_{i}; \phi)
  \end{array}
\label{eq:iwaeelbograd}
\end{equation}

In this paper, we consider the problem of training NIC with multiple-sample IWAE target (Eq.~\ref{eq:iwaeelbo}), which allows us to learn a richer latent space. First, we recognise that NIC's factorized uniform variational posterior has impacts on variance and bias properties of gradient estimators. Specifically, we find NIC's pathwise gradient estimator equivalent to an improved STL estimator \citep{roeder2017sticking}, which is unbiased even for the IWAE target. However, NIC's IWAE-DReG estimator \citep{tucker2018doubly} has extra bias, which causes performance decay. Moreover, we provide insights on a commonly adopted but little explained trick of training NIC from gradient variance perspective. Based on those analysis and observations, we further propose MS-NIC, a novel improvement of multiple-sample IWAE target for NIC. Experimental results show that it improves sota NIC methods \citep{balle2018variational, cheng2020learned} and learns richer latent representation. Our method is plug-and-play, and can be extended into neural video compression.

To wrap up, our contributions are as follows:
\begin{itemize}
    \item We provide insights on the impact of the uniform variational posterior upon gradient estimators, and a commonly adopted but little discussed trick of NIC training from gradient variance perspective.
    \item We propose multiple-sample neural image compression (MS-NIC). It is a novel enhancement of hierarchical IWAE \citep{burda2016importance} for neural image compression. To the best of our knowledge, we are the first to consider a tighter ELBO for training neural image compression.
    \item We demonstrate the efficiency of MS-NIC through experimental results on sota NIC methods. Our method is plug-and-play for neural image compression and can be easily applied to neural video compression.

\end{itemize}

\section{Gradient Estimation for Neural Image Compression}
The common NIC framework (Eq.~\ref{eq:gg17elbo}, Eq~\ref{eq:gg18elbo}) adopts fully factorized uniform distribution $q(\bmt{y}, \bmt{z}|\bm{x}) = \prod \mathcal{U}(y^i-\frac{1}{2}, y^i+\frac{1}{2})\prod \mathcal{U}(z^j-\frac{1}{2}, z^j+\frac{1}{2})$ to simulate the quantization noise. Such formulation has the following special properties:
\begin{itemize}
    \item Property I: $q(\bmt{z}|\bmt{y})$ and $q(\bmt{y}|\bm{x})$'s support depends on the parameter.
    \item Property II: $\log q(\bmt{z}|\bmt{y}) = \log q(\bmt{y}|\bm{x}) = 0$ on their support.
\end{itemize}

The impacts of these two properties are frequently neglected in previous works, which does not influence the results for single-sample pathwise gradient estimators (a.k.a. reparameterization trick in \citet{kingma2013auto}). In this section, we discuss the impacts of these two properties upon the variance and biasness of gradient estimators. Our analysis is based on single level latent (Eq.~\ref{eq:gg17elbo}) instead of hierarchical latent (Eq.~\ref{eq:gg18elbo}) to simplify notations.

\subsection{Impact on Pathwise Gradient Estimators}
\label{section:iopge}

First, let's consider the single-sample case. We can expand the pathwise gradient of ELBO in Eq.~\ref{eq:gg17elbograd} into Eq.~\ref{eq:gg17elboexp}. As indicated in the equation, $\phi$ contributes to $\mathcal{L}$ in two ways. The first way is through the reparametrized $\bm{\tilde{y}}(\bm{\epsilon}; \phi)$ (pathwise term), and the other way is through the parameter of $\log q(\bmt{y}|\bm{x})$ (parameter score term). Generally, the parameter score term has higher variance than the pathwise term. The STL \citep{roeder2017sticking} reduces the gradient by dropping the score. It is unbiased since the dropped term's expectation $\mathbb{E}_{q(\bmt{y}|\bm{x})}[\nabla_{\phi}\log q_{\phi}(\bmt{y}|\bm{x})]$ is $0$. 

\begin{equation}
  \begin{array}{l}
  \nabla_{\phi} \mathcal{L} = \mathbb{E}_{p(\bm{\epsilon})}[\underbrace{\nabla_{\bmt{y}}(\log \frac{p(\bm{x}|\bmt{y}) p(\bmt{y})}{q(\bmt{y}|\bm{x})})\nabla_{\phi}\bmt{y}(\bm{\epsilon}; \phi)}_{\textrm{\scriptsize{pathwise term}}} - \underbrace{\nabla_{\phi}\log q_{\phi}(\bmt{y}|\bm{x})}_{\textrm{\scriptsize{parameter score term}}}]
  \end{array}
\label{eq:gg17elboexp}
\end{equation}


Now let's consider the STL estimator of multiple-sample IWAE bound (Eq.~\ref{eq:iwaeelbo}). As shown in \citet{tucker2018doubly}, the STL estimation of IWAE bound gradient is biased. 
To reveal the reason, consider expanding the gradient Eq.~\ref{eq:iwaeelbograd} into partial derivatives as we expand Eq.~\ref{eq:gg17elbograd} into Eq.~\ref{eq:gg17elboexp}. Unlike single-sample case, the dropped parameter score term $\mathbb{E}_{p(\bm{\epsilon}_{1:k})}[\sum \tilde{w}_{i} (- \nabla_{\phi}\log q_{\phi}(\bmt{y}|\bm{x}))]$ is no longer $0$ due to the importance weight $\tilde{w}_{i}$. This means that STL loses its unbiasness in general IWAE cases.
 
Regarding NIC, however, the direct pathwise gradient for IWAE bound is automatically an unbiased STL estimator. Property II means that variational posterior has constant entropy, which further means that the parameter score gradient is $0$. So, NIC's pathwise gradient of IWAE bound is equvailent to an extended, unbiased STL estimator.

\subsection{Impact on Score Function Gradient Estimators}
\label{section:iosfge}

In previous section, we show the bless of NIC's special properties on pathwise gradient estimators. In this section, we show their curse on score function gradient estimators. Sepcifically, Property I implies that $q(\bm{\tilde{z}}|\bm{\tilde{y}})$ and $q(\bm{\tilde{y}}|\bm{\tilde{x}})$ are not absolute continuous, and hence the score function gradient estimators of those distributions are biased.

For example, consider a univariate random variable $x \sim p_{\theta}(x) = \mathcal{U}(\theta-\frac{1}{2}, \theta+\frac{1}{2})$. Our task is to estimate the gradient of a differentiable function $f(x)$. And consider the $\theta$-independent random variable $\epsilon \sim p(\epsilon) = \mathcal{U}(-\frac{1}{2}, +\frac{1}{2})$, the transform $x(\epsilon; \theta) = \theta + \epsilon$. Under such conditions, the Monte Carlo estimated pathwise gradient and score function gradient are:
\begin{equation}
  \begin{array}{l}
        \textrm{pathwise gradient: } \nabla_{\theta} \mathbb{E}_{p_{\theta}(x)} [f(x)] = \nabla_{\theta} \mathbb{E}_{p(\epsilon)}[f(x(\epsilon; \theta))] \approx \frac{1}{N}\overset{N}{\underset{i}{\sum}} \nabla_{\theta} f(\theta + \epsilon_{i})
  \end{array}
\label{eq:pw_example}
\end{equation}
\begin{equation}
  \begin{array}{l}
       \textrm{score function gradient: } \nabla_{\theta} \mathbb{E}_{p_{\theta}(x)} [f(x)] =  \mathbb{E}_{p_{\theta}(x)}[\nabla_{\theta} \log p_{\theta}(x)f(x)] = 0
  \end{array}
\label{eq:sf_example}
\end{equation}

Eq.~\ref{eq:pw_example} does not equal to Eq.~\ref{eq:sf_example}, and Eq.\ref{eq:sf_example} is wrong. The score function gradient is only unbiased when the distribution satisfies the absolute continuity condition of \citep{mohamed2020monte}. This reflects that under the formulation of NIC, the equivalence between the score function gradient (a.k.a. REINFORCE \citep{williams1992simple}) and pathwise gradient (a.k.a reparameterization trick in \citep{kingma2013auto}) no longer holds.

\begin{table}[ht]
\centering
\caption{Effect of DReG gradient estimator in NIC.}
\vspace{2mm}
\label{tab:dreg}
\begin{tabular}{lccccc}
\toprule
                      & Sample Size & bpp & MSE & PSNR (db) & R-D cost          \\ \midrule 
\textit{Single-sample} &&&&& \\
Baseline \citep{balle2018variational} & - & 0.5273 & 32.61 & 33.28 & 1.017 \\ \midrule
\textit{Multiple-sample} &&&&& \\
MS-NIC-MIX(pathwise gradient)     & 5 & 0.5259 & 31.84 & 33.38 & 1.003 \\
MS-NIC-MIX(DReG gradient)         & 5 & 0.5316 & 35.09 & 32.90 & 1.058 \\ \bottomrule
\end{tabular}
\end{table}
Such equivalence is the cornerstone of many gradient estimators, and IWAE-DReG \citep{tucker2018doubly} is one of them.  IWAE-DReG is a popular gradient estimator for IWAE target (Eq.~\ref{eq:iwaeelbo}) as it resolves the vanish of inference network gradient SNR (signal to noise ratio). However, the correctness of IWAE-DReG depends on the equivalence between the score function gradient and pathwise gradient, which does not hold for NIC. Specifically, IWAE-DReG expand the total derivative of IWAE target as Eq.~\ref{eq:dreg1} and perform another round of reparameterization on the score function term as Eq.~\ref{eq:dreg2} to further reduce the gradient variance. However, Eq.~\ref{eq:dreg2} requires the equivalence of pathwise gradient and score function gradient.
\begin{align}
\nabla_{\phi} \mathbb{E}_{q_{\phi}(\bmt{y}_{1:k}|\bm{x})}[\log \frac{1}{k}\sum_{i=1}^{k}w_i]= \mathbb{E}_{p(\bm{\epsilon}_{1:k})}[\sum_{i=1}^{k}\underbrace{\frac{w_i}{\sum_{j=1}^{k}w_j}(-\frac{\partial \log q_{\phi}(\bmt{y}_i|\bm{x})}{\partial \phi}}_{\textrm{score function term}}+\frac{\partial \log  w(\bm{\epsilon}_i;\phi)}{\partial \bmt{y}_i}\frac{\partial \bmt{ y}(\bm{\epsilon}_i;\phi)}{\partial \phi})]
\label{eq:dreg1}
\end{align}
\begin{align}
\mathbb{E}_{q(\bmt{y}_i|\bm{x})}[\frac{w_i}{\sum_{j=1}^{k}w_j}\frac{\partial \log q_{\phi}(\bmt{y}_i|\bm{x})}{\partial \phi}]= \mathbb{E}_{p(\bm{\epsilon_i})}[\frac{\partial}{\partial \bmt{y}_i}(\frac{w_i}{\sum_{j=1}^{k}w_j})\frac{\partial \bmt{y}(\bm{\epsilon}_i;\phi)}{\partial \phi_i}]
\label{eq:dreg2}
\end{align}
As we show empirically in Tab.~\ref{tab:dreg}, blindly adopting IWAE-DReG estimator for multiple-sample NIC brings evident performance decay. Other than IWAE-DReG, many other graident estimators such as NVIL \citep{mnih2014neural}, VIMCO \citep{mnih2016variational} and GDReG \citep{bauer2021generalized} do not apply to NIC. They either bring some extra bias or are totally wrong.

\subsection{The \textit{direct-y} Trick in Training NIC}
In NIC, we feed deterministic parameter $\bm{y}$ into z inference model instead of noisy samples $\bmt{y}$. This implies that $\bmt{z}$ is sampled from $q(\bmt{z}|\bm{y})$ instead of $q(\bmt{z}|\bmt{y})$. This trick is initially adopted in \citet{balle2018variational} and followed by most of the subsequent works. However, it is little discussed. In this paper, we refer it to \textit{direct-y} trick. \citet{yang2020improving} observes that feeding $\bmt{y}$ instead of $\bm{y}$ causes severe performance decay. We confirm this result in Tab.~\ref{tab:directy}. Thus, \textit{direct-y} trick is essential to train hierarchical NIC. 

\begin{table}[ht]
\centering
\caption{Effects of \textit{direct-y} on R-D performance. 2-level VAE is equivalent to \citet{balle2018variational} without \textit{direct-y}.}
\vspace{2mm}
\label{tab:directy}
\begin{tabular}{lcccc}
\toprule
                      & bpp & MSE & PSNR & R-D cost          \\ \midrule
2-level VAE & 0.9968 & 33.08 & 33.22 & 1.493 \\
\citep{balle2018variational} & 0.5273 & 32.61 & 33.28 & 1.017 \\ \bottomrule
\end{tabular}
\end{table}
\begin{table}[ht]
\centering
\caption{Effects of \textit{direct-y} on gradient SNR of different parts of the model. 2-level VAE is equivalent to \citet{balle2018variational} without \textit{direct-y}. "early" is $5 \times 10^4$ iterations, "mid" is $5 \times 10^5$ iterations and "late" is $1 \times 10^6$ iterations.  "infer" is the abbreviation for "inference model", and "gen" is the abbreviation for "generative model".}
\vspace{2mm}
\label{tab:directysnr}
\begin{tabular}{llccccc}
\toprule
                     & & \multicolumn{5}{c}{gradient SNR of \#} \\ \cmidrule(l){3-7}
                      Iteration & Method & y infer & y gen & z infer & z gen & z prior          \\ \midrule
early & 2-level VAE & 2.287 & 0.5343 & 0.3419 & 0.4099 & 0.9991 \\
 & \citet{balle2018variational} & 2.174 & 0.5179 & 0.5341 & 0.3813 & 1.069 \\ \midrule
mid & 2-level VAE & 1.350 & 0.4793 & 0.2414 & 0.3583 & 0.8861 \\
 & \citet{balle2018variational} & 1.334 & 0.4813 & 0.4879 & 0.3761 & 0.9693 \\ \midrule
late & 2-level VAE & 1.217 & 0.4746 & 0.2863 & 0.3439 & 0.8691 \\
 & \citet{balle2018variational} & 1.206 & 0.4763 & 0.5506 & 0.3707 & 0.9339 \\ \bottomrule
\end{tabular}
\end{table}

One explanation is to view $q(\bmt{z}|\bm{y})$ as $q(\bmt{z}|\bm{x})$, and $q(\bmt{y}, \bmt{z}|\bmt{x})$ factorized as $q(\bmt{y}|\bm{x})q(\bmt{z}|\bm{x})$ (See Fig.\ref{fig:plate} (a)-(c)). A similar trick of feeding mean parameter can be traced back to the Helmholtz machine \citep{dayan1995helmholtz}. However, this provides a rationale why \textit{direct-y} is fine to be adopted but does not explain why samping $\bmt{z}$ from $q(\bmt{z}|\bmt{y})$ fails. We provide an alternative explanation from the gradient variance perspective. Specifically, $q(\bmt{z}|\bmt{y})$ has two stochastic arguments that could cause high variance in the gradient of z inference model, and make its convergence difficult. To verify this, we follow \citet{rainforth2018tighter} to compare the gradient SNR, which is the absolute value of the empirical mean divided by standard deviation. We trace the gradient SNR during different training stages as model converges (See Sec.~\ref{sec:eset} for detailed setups).

As demonstrated in Tab.~\ref{tab:directysnr}, the gradient SNR of z inference model of standard 2-level VAE (without \textit{direct y}) is indeed significantly lower than \citet{balle2018variational} (with \textit{direct y}) during all 3 stage of training. This result reveals that the z inference model is more difficult to train without $\textit{direct-y}$. And such difficulty could be the source of the failure of NIC without \textit{direct-y} trick.

\section{Multiple-sample Neural Image Compression}
In this section, we consider the multiple-sample approach based on the 2-level hierarchical framework by \citet{balle2018variational}, which is the de facto NIC architecture adopted by many sota methods. To simplify notations, $\log q(\bmt{z}|\bmt{y})$ and $\log q(\bmt{y}|\bm{x})$ in ELBO are omitted as they are $0$.
\begin{figure}[thb]
\begin{center}
    \includegraphics[width=0.8\linewidth]{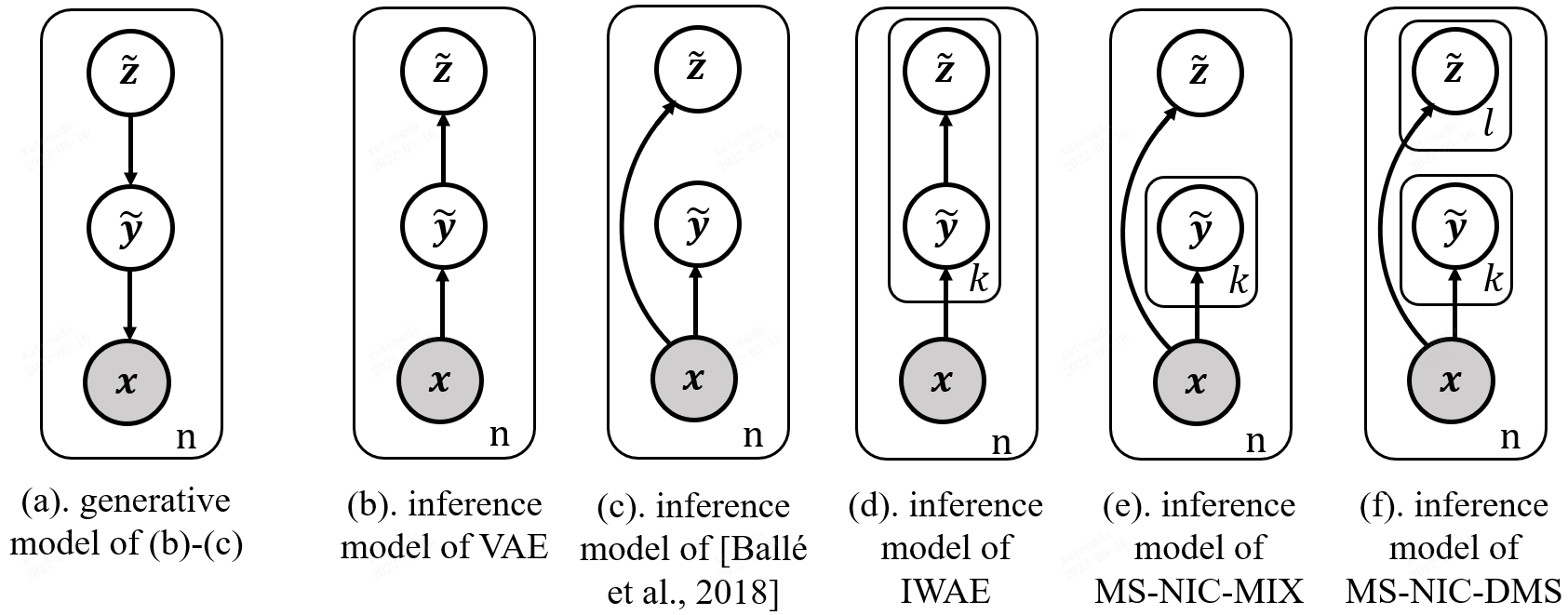}
    \caption{The plate notation of different NIC methods. $\bm{x}$ is the observed image, $\bmt{y}$ and $\bmt{z}$ are latent. The inference models show how we sample from variational posterior duing training. $n$ is the number of data points in dataset, $k, l$ is the sample size of multiple-sample approaches. The generative model of (b), (c) is (a). The generative model of (d)-(f) is shown in Appendix.~\ref{sec:plate2}. For clarity, we omit the parameters.}
    \label{fig:plate}
\end{center}
\end{figure}
First, let's consider directly applying 2-level IWAE to NIC without \textit{direct-y} trick (See Fig.~\ref{fig:plate} (d)). Regarding a $k$ sample IWAE, we first compute parameter $\bm{y}$ of $q(\bmt{y}|\bm{x})$ and sample $\bmt{y}_{1:k}$ from it. Then, we compute parameter $\bm{z}_{1:k}$ of $q(\bmt{z}_{1:k}|\bmt{y}_{1:k})$ and samples $\bmt{z}_{1:k}$ from it. Afterward, $\bmt{y}_{1:k}$ and $\bmt{z}_{1:k}$ are fed into the generative model and compute $w_{1:k}$. Finally, we follow Eq~\ref{eq:iwaeelbograd} to compute the gradient and update parameters. In fact, this is the standard 2-level IWAE in the original IWAE paper.

However, the vanilla 2-level IWAE becomes a problem for NIC with \textit{direct-y} trick. Concerning a $k$ sample IWAE, we sample $\bmt{y}_{1:k}$ from $q(\bmt{y}|x)$. Due to the \textit{direct-y} trick, we feed $\bm{y}$ instead of $\bmt{y}_{1:k}$ into z inference network, and our $q(\bmt{z}|\bm{y})$ has only one parameter $\bm{z}$ other than $k$ parameter $\bm{z}_{1:k}$. If we follow the 2-level IWAE approach, only one sample $\bmt{z}$ is obtained, and $w_{1:k}$ can not be computed. One method is to limit the multiple-sample part to $\bmt{y}$ related term only and optimize other parts via single-sample SGVB-1, which produces our MS-NIC-MIX (See Fig~\ref{fig:plate} (e)). Another method is to sample another $l$ samples of $\bmt{z}_j$ from $q(\bmt{z}|\bm{y})$ and nest it with MS-NIC-MIX, which generates our MS-NIC-DMS (See Fig~\ref{fig:plate} (f)).

\subsection{MS-NIC-MIX: Multiple-sample NIC with Mixture}

One way to optimize multiple-sample IWAE target of NIC with \textit{direct-y} trick is to sample $\bmt{y}$ $k$ times to obtain $\bmt{y}_{1:k}$ and $\bmt{z}$ only $1$ time. Then we perform $k$ sample log mean of $p(\bm{x}|\bmt{y}_i)p(\bmt{y}_i|\bmt{z})$ to obtain a multiple-sample estimated $\log p(\bm{x}|\bmt{z})$, add it with single-sample $\log p(\bmt{z})$. This brings a $\mathcal{L}^{MIX}_{k}$ with the form of a mixture of 1-level VAE and 1-level IWAE ELBO:

\begin{equation}
  \begin{array}{l}
  \mathcal{L}_{k}^{MIX} = \mathbb{E}_{q_{\phi}(\bmt{z}|\bm{x})}[\mathbb{E}_{q_{\phi}(\bmt{y}_{1:k}|\bm{x})}[\log \frac{1}{k}\overset{k}{\underset{i}{\sum}} p(\bm{x}|\bmt{y}_{i})p(\bmt{y}_{i}|\bmt{z})|\bmt{z}] + \log p(\bmt{z})]
  \end{array}
\label{eq:yelbo}
\end{equation}

Moreover, $\mathcal{L}^{MIX}_k$ is a reasonably preferable target over ELBO as it satisfies the following properties (See Appendix.~\ref{app:pf} for proof):

\begin{enumerate}
    \item $\log p(x) \ge \mathcal{L}^{MIX}_k $
    \item $\mathcal{L}^{MIX}_k\ge  \mathcal{L}^{MIX}_m$ for $k \ge m$
\end{enumerate}

Although $\mathcal{L}^{MIX}_k$ does not converge to true $\log p(\bm{x})$ as $k$ grows, it is still a lower bound of $\log p(\bm{x})$ and tighter than ELBO (as $\mathcal{L}^{MIX}_1 = $ ELBO). Its gradient can be computed via pathwise estimator. Denote the per-sample integrand $p(\bm{x}|\bmt{y}_i)p(\bmt{y_i}|\bmt{z})$ as $w^{MIX}_i$, and its relative weight as $\tilde{w}^{MIX}_i$, then the gradient $\nabla_{\phi}\mathcal{L}_{k}^{MIX}$ can be estimated as Eq.~\ref{eq:yelbograd}.

\begin{equation}
  \begin{array}{l}
  \mathcal{L}_{k}^{MIX} =  \mathbb{E}_{p(\bm{\epsilon}^{y}_{1:k}, \bm{\epsilon}^{z})}[\log \frac{1}{k}\overset{k}{\underset{i}{\sum}} p(\bm{x}|\bmt{y}(\bm{\epsilon}^{y}_{i}; \phi))p(\bmt{y}(\bm{\epsilon}^{y}_{i}; \phi)|\bmt{z}(\bm{\epsilon}^{z}; \phi)) + \log p(\bmt{z}(\bm{\epsilon}^{z}; \phi))] \\ 
  \hspace{3em} \approx \log \frac{1}{k}\overset{k}{\underset{i}{\sum}} \underbrace{p(\bm{x}|\bmt{y}(\bm{\epsilon}^{y}_{i}; \phi))p(\bmt{y}(\bm{\epsilon}^{y}_{i}; \phi)|\bmt{z}(\bm{\epsilon}^{z}; \phi))}_{w^{MIX}(\bm{\epsilon}_{1:k}^{y}, \bm{\epsilon}^{z}; \phi)} + \log p(\bmt{z}(\bm{\epsilon}^{z}; \phi))
  
  \end{array}
\label{eq:yelbort}
\end{equation}

\begin{equation}
  \begin{array}{l}
  \nabla_{\phi}\mathcal{L}_{k}^{MIX} \approx \overset{k}{\underset{i}{\sum}} \tilde{w}_{i}^{MIX} \nabla_{\phi} \log w^{MIX}(\bm{\epsilon}_{1:k}^{y}, \bm{\epsilon}^{z}; \phi) + \nabla_{\phi} \log p(\bmt{z}(\bm{\epsilon}^{z}; \phi))
  \end{array}
\label{eq:yelbograd}
\end{equation}

Another way to understand MS-NIC-MIX is to view the y inference/generative model as a single level IWAE, and the z inference/generative model as a large prior of $\bmt{y}$ which is optimized via SGVB-1. This perspective is often taken by works in NIC context model \citep{minnen2018joint, he2021checkerboard}, as the context model of NIC is often limited to $\bmt{y}$.

\subsection{MS-NIC-DMS: Multiple-sample NIC with Double Multiple Sampling}

An intuitive improvement over MS-NIC-MIX is to add another round of multiple-sample over $\bmt{z}$. Specifically, we sample $\bmt{z}$ $l$ times, nest it with $\mathcal{L}^{MIX}_{k}$ to obtain $\mathcal{L}_{k,l}^{DMS}$: 

\begin{equation}
  \begin{array}{l}
  \mathcal{L}_{k,l}^{DMS} = \mathbb{E}_{q_{\phi}(\bmt{z}_{1:l}|\bm{x})}[\log \frac{1}{l}\overset{l}{\underset{j}{\sum}} \exp{(\mathbb{E}_{q_{\phi}(\bmt{y}_{1:k}|\bm{x})}[\log \frac{1}{k}\overset{k}{\underset{i}{\sum}} p(\bm{x}|\bmt{y}_{i})p(\bmt{y}_{i}|\bmt{z}_{j})|\bmt{z}_{j}])}p(\bmt{z}_{j})] 
  \end{array}
\label{eq:yzelbo18hat}
\end{equation}

And we name it MS-NIC-DMS as it adopts multiple sampling twice. Moreover, $\mathcal{L}_{k,l}^{DMS}$ is a reasonably better target for optimizaion over ELBO and $\mathcal{L}_{k}^{MIX}$, as it satisfies the following properties (See proof in Appendix.~\ref{app:pf}):

\begin{enumerate}
    \item $\log p(\bm{x}) \ge \mathcal{L}^{DMS}_{k, l} $
    \item $\mathcal{L}^{DMS}_{k, l}\ge  \mathcal{L}^{DMS}_{m, n}$ for $k \ge m, l \ge n$
    \item $\mathcal{L}^{DMS}_{k, l} \ge \mathcal{L}^{MIX}_{k}$
    \item $\mathcal{L}^{DMS}_{k, l} \rightarrow \log p(\bm{x})$ as $k, l \rightarrow \infty$, under the assumption that $\log \frac{p(\bm{x}|\bmt{y}_{i})p(\bmt{y}_{i}|\bmt{z}_{j})}{q(\bmt{y}_i|\bm{x})}$ and $\log \frac{p(\bm{x}|\bmt{z}_{j})p(\bmt{z}_{j})}{q(\bmt{z}_j|\bm{x})}$ are bounded.
\end{enumerate}

In other words, the target $\mathcal{L}_{k,l}^{DMS}$ is a lowerbound of $\log p(\bm{x})$, converging to $\log p(\bm{x})$ as $k, l \rightarrow \infty$, tighter than $\mathcal{L}_{k}^{MIX}$ and tighter than ELBO (as $\mathcal{L}^{DMS}_{1,1} = $ ELBO). However, its Monte Carlo estimation is biased due to the nested transformation and expectation. Empirically, we find that directly adopting biased 
pathwise estimator works fine. And its gradient can be estimated by pathwise estimator similar to original IWAE target (See Eq.~\ref{eq:iwaeelbograd}). 

\begin{equation}
  \begin{array}{l}
  \mathcal{L}_{k, l}^{DMS} = \mathbb{E}_{q( \bm{\epsilon}^{z}_{1:l})}[\log \frac{1}{l}\overset{l}{\underset{j}{\sum}}\exp (\mathbb{E}_{q(\bm{\epsilon}^{y}_{k:l})}[\log \frac{1}{k}\overset{k}{\underset{i}{\sum}}
  p(\bm{x}|\bmt{y}(\bm{\epsilon}^{y}_{i}; \phi))p(\bmt{y}(\bm{\epsilon}^{y}_{i}; \phi)|\bmt{z}(\bm{\epsilon}^{z}_{j}; \phi))])p(\bmt{z}(\bm{\epsilon}^{z}_{j}; \phi))] \\
 \hspace{3em} \approx \log \frac{1}{l}\overset{l}{\underset{j}{\sum}} \frac{1}{k}\overset{k}{\underset{i}{\sum}}
 \underbrace{
  p(\bm{x}|\bmt{y}(\bm{\epsilon}^{y}_{i}; \phi))p(\bmt{y}(\bm{\epsilon}^{y}_{i}; \phi)|\bmt{z}(\bm{\epsilon}^{z}_{j}; \phi))p(\bmt{z}(\bm{\epsilon}^{z}_{j}; \phi))}_{w^{DMS}(\bm{\epsilon}_{1:k}^{y}, \bm{\epsilon}_{1:l}^{z}; \phi)}
  \end{array}
\label{eq:yzelbo}
\end{equation}

Another interpretation of MS-NIC-DMS is to view it as a multiple level IWAE with repeated local samples. The $\mathcal{L}_{k, l}^{DMS}$ Monte Carlo pathwise estimator has the form of IWAE with $k \times l$ samples. However, there are multiple repeated samples that contain the same $\bmt{y}_i$ and $\bmt{z}_j$. For example, the samples $w^{IWAE}_{1:6}$ of 2 level IWAE with sample size 6 look like Eq.~\ref{eq:iwae6}. While the samples $w^{DMS}_{1:2, 1:3}$ of MS-NIC-DMS with $2 \times 3$ samples look like Eq.~\ref{eq:dms6}. We can see that in IWAE, we have 6 pairs of independently sampled $\bmt{y}$ and $\bmt{z}$, while in MS-NIC-DMS, we have 2 independent $\bmt{y}$ and 3 independent $\bmt{z}$, they are paired to generate 6 samples in total. Note that this is only applicable to NIC as $\bmt{y}$ and $\bmt{z}$ are conditionally independent given $\bmt{x}$ due to \textit{direct-y} trick.

\noindent\begin{minipage}{.5\linewidth}
\begin{equation}
  \begin{array}{l}
    {w^{IWAE}_{1:6}} = \{p(\bm{x}|\bmt{y}_1)p(\bmt{y}_1|\bmt{z}_1)p(\bmt{z}_1), \\
    \hspace{5em} p(\bm{x}|\bmt{y}_2)p(\bmt{y}_2|\bmt{z}_2)p(\bmt{z}_2), \\
    \hspace{5em} p(\bm{x}|\bmt{y}_3)p(\bmt{y}_3|\bmt{z}_3)p(\bmt{z}_3), \\
    \hspace{5em} p(\bm{x}|\bmt{y}_4)p(\bmt{y}_4|\bmt{z}_4)p(\bmt{z}_4), \\
    \hspace{5em} p(\bm{x}|\bmt{y}_5)p(\bmt{y}_5|\bmt{z}_5)p(\bmt{z}_5), \\ 
    \hspace{5em} p(\bm{x}|\bmt{y}_6)p(\bmt{y}_6|\bmt{z}_6)p(\bmt{z}_6)\}
  \end{array}
\label{eq:iwae6}
\end{equation}
\end{minipage}%
\begin{minipage}{.5\linewidth}
\begin{equation}
  \begin{array}{l}
    {w^{DMS}_{1:2, 1:3}} = \{ p(\bm{x}|\bmt{y}_1)p(\bmt{y}_1|\bmt{z}_1)p(\bmt{z}_1), \\
    \hspace{4.8em} p(\bm{x}|\bmt{y}_1)p(\bmt{y}_1|\bmt{z}_2)p(\bmt{z}_2), \\
    \hspace{4.8em} p(\bm{x}|\bmt{y}_1)p(\bmt{y}_1|\bmt{z}_3)p(\bmt{z}_3), \\
    \hspace{4.8em} p(\bm{x}|\bmt{y}_2)p(\bmt{y}_2|\bmt{z}_1)p(\bmt{z}_1), \\
    \hspace{4.8em} p(\bm{x}|\bmt{y}_2)p(\bmt{y}_2|\bmt{z}_2)p(\bmt{z}_2), \\ 
    \hspace{4.8em} p(\bm{x}|\bmt{y}_2)p(\bmt{y}_2|\bmt{z}_3)p(\bmt{z}_3) \}
  \end{array}
\label{eq:dms6}
\end{equation}
\end{minipage}



\section{Related Work}
\subsection{Lossy Neural Image and Video Compression}

\citet{balle2017end} and \citet{balle2018variational} formulate lossy neural image compression as a variational inference problem, by interpreting the additive uniform noise (AUN) relaxed scalar quantization as a factorized uniform variational posterior. After that, the majority of sota lossy neural image compression methods adopt this formulation \citep{minnen2018joint, minnen2020channel, cheng2020learned, guo2021causal, gao2021neural, he2022elic}. And \citet{yang2020improving, guo2021soft} also require a AUN trained NIC as base. Moreover, the majority of neural video compression also adopts this formulation \citep{lu2019dvc, lu2020end, agustsson2020scale, hu2021fvc, li2021deep}, implying that MS-NIC can be extended to video compression without much pain.

Other approaches to train NIC include random rounding \citep{toderici2015variable, toderici2017full} and straight through estimator (STE) \citep{theis2017lossy}. Another promising approach is the VQ-VAE \citep{van2017neural}. By the submission of this manuscript, one unarchived work \citep{zhu2022unified} has shown the potential of VQ-VAE in practical NIC. Our MS-NIC does not apply to the approaches mentioned in this paragraph, as the formulation of variational posterior is different.

\subsection{Tighter Lower Bound for VAE}

IWAE \citep{burda2016importance} stirs up the discussion of adopting tighter lower bound for training VAEs. However, at the first glance it is not straightforward why it might works. \citet{cremer2018inference} decomposes the inference suboptimality of VAE into two parts: 1) The limited expressiveness of interence model. 2) The gap between ELBO and log likelihood. However, this gap refers to inference not training. The original IWAE paper empirically shows that IWAE can learn a richer latent representation. And \citet{cremer2017reinterpreting} shows that the IWAE target converges to ELBO under the expectation of true posterior. And thus the posterior collapse is avoided.

From the information preference \citep{chen2017variational} perspective, VAE prefers to distribute information in generative distribution than autoencoding information in the latent. This preference formulates another view of posterior collapse. And it stems from the gap between ELBO and true log likelihood. There are various approaches alleviating it, including \textit{soft free bits} \citep{theis2017lossy} and \textit{KL annealing} \citep{serban2017hierarchical}. In our opinion, IWAE also belongs to those methods, and it is asymptotically optimal. However, we have not found many works comparing IWAE with those methods. Moreover, those approaches are rarely adopted in NIC community. 

Many follow-ups of IWAE stress gradient variance reduction \citep{roeder2017sticking, tucker2018doubly, rainforth2018tighter}, discrete latent \citep{mnih2016variational} and debiasing IWAE target \citep{nowozin2018debiasing}. Although the idea of tighter low bound training has been applied to the field of neural joint source channel coding \citep{choi2018necst, song2020infomax}, to the best of our knowledge, no work in NIC consider it yet.
   
\subsection{Multi-Sample Inference for Neural Image Compression}

\citet{theis2021importance} considers the similar topic of importance weighted NIC. However, it does not consider training of NIC. Instead, it focuses on achieving IWAE target with an entropy coding technique named \textit{softmin}, just like BB-ANS \citep{townsend2018practical} achieving ELBO. It is alluring to apply \textit{softmin} to MS-NIC, as it closes the multiple-sample training and inference gap. However, it requires large number of samples (e.g. $4096$) to achieve slight improvement for 64$\times$64 images. The potential sample size required for practical NIC is forbidding. Moreover, we believe the stochastic lossy encoding scheme \citep{agustsson2020universally} that \citet{theis2021importance} is not yet ready to be applied (See Appendix.~\ref{app:teit} for details).

\section{Experimental Results}

\subsection{Experimental Settings}
\label{sec:eset}

Following \citet{he2022elic}, we train all the models on the largest 8000 images of ImageNet \citep{deng2009imagenet}, followed by a downsampling according to \citet{balle2018variational}. And we use Kodak \citep{kodak} for evaluation. For the experiments based on \citet{balle2018variational} (include Tab.~\ref{tab:dreg}, Tab.~\ref{tab:directy}), we follows the setting of the original paper except for the selection of $\lambda$s, For the selection of $\lambda$s, we set $\lambda \in \{0.0016, 0.0032, 0.0075, 0.015, 0.03, 0.045, 0.08\}$ as suggested in \citet{cheng2020learned}. And for the experiments based on \citet{cheng2020learned}, we follows the setting of original paper. More detailed experimental settings can be found in Appendix.~\ref{ap:detailsset}.

And when comparing the R-D performance of models trained on multiple $\lambda$s, we use Bjontegaard metric (BD-Metric) and Bjontegaard bitrate (BD-BR) \citep{bjontegaard2001calculation}, which is widely applied when comparing codecs. More detailed experimental results can be found in Appendix.~\ref{ap:detailsexp}.

\begin{table}[ht]
\centering
\caption{Results based on \citet{balle2018variational}.}
\vspace{2mm}
\label{tab:gg18results}
\begin{tabular}{lcccc}
\toprule
                     & \multicolumn{2}{c}{PSNR} & \multicolumn{2}{c}{MS-SSIM} \\ \cmidrule(l){2-3} \cmidrule(l){4-5}
                      & BD-BR (\%) & BD-Metric & BD-BR (\%) & BD-Metric \\ \midrule
\textit{Single-sample} && \\
Baseline \citep{balle2018variational} & 0.000 & 0.000 & 0.000 & 0.0000\\ \midrule
\textit{Multiple-sample} && && \\
IWAE \citep{burda2016importance}   & 64.23 & -2.318 & 68.67 & -0.01648  \\
MS-NIC-MIX      & -3.847 & 0.1877 & -4.743 & 0.001618 \\
MS-NIC-DMS      & -4.929 & 0.2405 & -5.617 & 0.001976 \\ \bottomrule

\end{tabular}
\end{table}

\begin{table}[ht]
\centering
\caption{Results based on \citet{cheng2020learned}. The BD Metrics of IWAE can not be computed as its R-D is not monotonously increasing.}
\vspace{2mm}
\label{tab:cheng20results}
\begin{tabular}{lcccc}
\toprule
                     & \multicolumn{2}{c}{PSNR} & \multicolumn{2}{c}{MS-SSIM} \\ \cmidrule(l){2-3} \cmidrule(l){4-5}
                      & BD-BR (\%) & BD-Metric & BD-BR (\%) & BD-Metric \\ \midrule
\textit{Single-sample} && \\
Baseline \citep{cheng2020learned} & 0.0000 & 0.0000 & 0.0000 & 0.0000\\ \midrule
\textit{Multiple-sample} && && \\
IWAE \citep{burda2016importance}   & - & - & - & -  \\
MS-NIC-MIX      & -1.852 & 0.0805 & 2.238 & -0.0006764 \\
MS-NIC-DMS      & -2.378 & 0.1046 & 1.998 & -0.0006054  \\ \bottomrule

\end{tabular}
\end{table}

\subsection{R-D Performance}
We evaluate the performance of MS-NIC-MIX and MS-NIC-DMS based on sota NIC methods \citep{balle2018variational, cheng2020learned}. Empirically, we find that MS-NIC-MIX works best with sample size 8, and MS-NIC-DMS with sample size 16. The experimental results on sample size selection can be found in Appendix.~\ref{ap:samplesize}. Without special mention, we set the sample size of MS-NIC-MIX to 8 and MS-NIC-DMS to 16. 

For \citet{balle2018variational}, MS-NIC-MIX saves around $4\%$ of bitrate compared with single-sample baseline (See Tab.~\ref{tab:gg18results}). And MS-NIC-DMS saves around $5\%$ of bitrate. On the other hand, the original IWAE suffers performance decay as it is not compatible with \textit{direct-y} trick. For \citet{cheng2020learned}, we find that both MS-NIC-MIX and NS-NIC-DMS suppress baseline in PSNR. However, it is not as evident as \citet{balle2018variational}. Moreover, the MS-SSIM is slightly lower than the baseline. This is probably due to the auto-regressive context model. Besides, the original IWAE without \textit{direct-y} trick suffers from severe performance decay in both cases. The BD metric of IWAE on \citet{cheng2020learned} can not be computed as its R-D is not monotonous increasing, we refer interested readers to Appendix.~\ref{ap:detailsexp} for details.

\subsection{Latent Space Representation of MS-NIC}

To better understand the latent learned by MS-NIC, we evaluate the variance and coefficient of variation (Cov) of per-dimension latent distribution mean parameter $\bm{y}^{(i)}, \bm{z}^{(i)}$, with regard to input distribution $p(\bm{x})$. As we are also interested in the discrete representation, we provide statistics of rounded mean $\bm{\bar{y}}^{(i)}, \bm{\bar{z}}^{(i)}$. These metrics show how much do latents vary when input changes, and a large variation in latents means that there are useful information encoded. A really small variation indicates that the latent is "dead" in that dimension. 

As shown in Tab.~\ref{tab:nord} of Appendix.~\ref{app:dlv}, the latent of multiple-sample approaches has higher variance than those of single-sample approach. Moreover, the Cov$(\bm{y})$ of multiple-sample approaches is around $4-5$ times higher than single-sample approach. Although the Cov$(\bm{z})$ of multiple-sample approaches is around $2$ times lower, the main contributor of image reconstruction is $\bm{y}$, and $\bm{z}$ only serves to predict $\bm{y}$'s distribution. Similar trend can be concluded from quantized latents $\bm{\bar{y}}, \bm{\bar{z}}$. From the variance and Cov perspective, the latent learned by MS-NIC is richer than single-sample approach. It is also noteworthy that although the variance and Cov of $\bm{y}, \bm{\bar{y}}$ of MS-NIC is significantly higher than single-sample approach, the bpp only varies slightly.

\begin{table}[ht]
\centering
\caption{The average of per-dimension latent variance and Cov across Kodak test images. The model is trained with $\lambda = 0.015$.}
\vspace{2mm}
\label{tab:per-dim}
\begin{tabular}{lcccccc}
\toprule
                     & \multicolumn{2}{c}{Var(\#)} & \multicolumn{2}{c}{Cov(\#)} & \multicolumn{2}{c}{bpp of \#} \\ \cmidrule(l){2-3} \cmidrule(l){4-5} \cmidrule(l){6-7}
                       Method & $\bm{\bar{y}}$ & $\bm{\bar{z}}$ & $\bm{\bar{y}}$ & $\bm{\bar{z}}$ & $\bm{\bar{y}}$ & $\bm{\bar{z}}$ \\ \midrule
\textit{Single-sample} & &  \\
\citet{balle2018variational} & 1.499 & 0.3255 & 19.70 & 9.944 & 0.5136 & 0.01342 \\ \midrule
\textit{Multiple-sample} &&&&&& \\
MS-NIC-MIX & 1.906 & 0.7594 & 111.1 & 7.425 & 0.5108 & 0.01521  \\
MS-NIC-DMS & 1.919 & 0.7648 & 95.51 & 7.243 & 0.5092 & 0.01634 \\\bottomrule
\end{tabular}
\end{table}

\section{Limitation \& Discussion}

A major limitation of our method is that the improvement in R-D performance is marginal, especially when based on \citet{cheng2020learned}. Moreover, evaluations on more recent sota methods are also helpful to strengthen the claims of this paper. In general, we think that the performance improvement of our approach is bounded by how severe the posterior collapse is in neural image compression. We measure the variance in latent dimension according to data in Fig.~\ref{fig:hst}. And from that figure it might be observed that the major divergence of IWAE and VAE happens when the variance is very small. And for the area where variance is reasonably large, the gain of IWAE is not that large. This probably indicates that the posterior collapse in neural image compression is only alleviated to a limited extend.

See more discussion in why the result on \citet{cheng2020learned} is negative in Appendix.~\ref{app:disc}

\section{Conclusion}
In this paper we propose MS-NIC, a multiple-sample importance weighted target for training NIC. It improves sota NIC methods and learns richer latent representation. A known limitation is that its R-D performance improvement is limited when applied to models with spatial context models (e.g. \citet{cheng2020learned}). Despite the somewhat negative result, this paper provides insights to the training of NIC models from VAE perspective. Further work could consider improving the performance and extend it into neural video compression.


\bibliography{ref_1}

\begin{thebibliography}{55}
\providecommand{\natexlab}[1]{#1}
\providecommand{\url}[1]{\texttt{#1}}
\expandafter\ifx\csname urlstyle\endcsname\relax
  \providecommand{\doi}[1]{doi: #1}\else
  \providecommand{\doi}{doi: \begingroup \urlstyle{rm}\Url}\fi

\bibitem[Agustsson and Theis(2020)]{agustsson2020universally}
E.~Agustsson and L.~Theis.
\newblock Universally quantized neural compression.
\newblock \emph{Advances in neural information processing systems},
  33:\penalty0 12367--12376, 2020.

\bibitem[Agustsson et~al.(2020)Agustsson, Minnen, Johnston, Balle, Hwang, and
  Toderici]{agustsson2020scale}
E.~Agustsson, D.~Minnen, N.~Johnston, J.~Balle, S.~J. Hwang, and G.~Toderici.
\newblock Scale-space flow for end-to-end optimized video compression.
\newblock In \emph{Proceedings of the IEEE/CVF Conference on Computer Vision
  and Pattern Recognition}, pages 8503--8512, 2020.

\bibitem[Ball{\'e} et~al.(2017)Ball{\'e}, Laparra, and
  Simoncelli]{balle2017end}
J.~Ball{\'e}, V.~Laparra, and E.~P. Simoncelli.
\newblock End-to-end optimized image compression.
\newblock In \emph{International Conference on Learning Representations}, 2017.

\bibitem[Ball{\'e} et~al.(2018)Ball{\'e}, Minnen, Singh, Hwang, and
  Johnston]{balle2018variational}
J.~Ball{\'e}, D.~Minnen, S.~Singh, S.~J. Hwang, and N.~Johnston.
\newblock Variational image compression with a scale hyperprior.
\newblock In \emph{International Conference on Learning Representations}, 2018.

\bibitem[Bauer and Mnih(2021)]{bauer2021generalized}
M.~Bauer and A.~Mnih.
\newblock Generalized doubly reparameterized gradient estimators.
\newblock In \emph{International Conference on Machine Learning}, pages
  738--747. PMLR, 2021.

\bibitem[Bjontegaard(2001)]{bjontegaard2001calculation}
G.~Bjontegaard.
\newblock Calculation of average psnr differences between rd-curves.
\newblock \emph{VCEG-M33}, 2001.

\bibitem[Bross et~al.(2021)Bross, Chen, Ohm, Sullivan, and
  Wang]{bross2021developments}
B.~Bross, J.~Chen, J.-R. Ohm, G.~J. Sullivan, and Y.-K. Wang.
\newblock Developments in international video coding standardization after avc,
  with an overview of versatile video coding (vvc).
\newblock \emph{Proceedings of the IEEE}, 109\penalty0 (9):\penalty0
  1463--1493, 2021.

\bibitem[Burda et~al.(2016)Burda, Grosse, and
  Salakhutdinov]{burda2016importance}
Y.~Burda, R.~B. Grosse, and R.~Salakhutdinov.
\newblock Importance weighted autoencoders.
\newblock In \emph{ICLR (Poster)}, 2016.

\bibitem[Chen et~al.(2017)Chen, Kingma, Salimans, Duan, Dhariwal, Schulman,
  Sutskever, and Abbeel]{chen2017variational}
X.~Chen, D.~P. Kingma, T.~Salimans, Y.~Duan, P.~Dhariwal, J.~Schulman,
  I.~Sutskever, and P.~Abbeel.
\newblock Variational lossy autoencoder.
\newblock In \emph{International Conference on Learning Representations}, 2017.

\bibitem[Cheng et~al.(2020)Cheng, Sun, Takeuchi, and Katto]{cheng2020learned}
Z.~Cheng, H.~Sun, M.~Takeuchi, and J.~Katto.
\newblock Learned image compression with discretized gaussian mixture
  likelihoods and attention modules.
\newblock In \emph{Proceedings of the IEEE/CVF Conference on Computer Vision
  and Pattern Recognition}, pages 7939--7948, 2020.

\bibitem[Choi et~al.(2018)Choi, Tatwawadi, Weissman, and Ermon]{choi2018necst}
K.~Choi, K.~Tatwawadi, T.~Weissman, and S.~Ermon.
\newblock Necst: neural joint source-channel coding.
\newblock 2018.

\bibitem[Cremer et~al.(2017)Cremer, Morris, and
  Duvenaud]{cremer2017reinterpreting}
C.~Cremer, Q.~Morris, and D.~Duvenaud.
\newblock Reinterpreting importance-weighted autoencoders.
\newblock 2017.

\bibitem[Cremer et~al.(2018)Cremer, Li, and Duvenaud]{cremer2018inference}
C.~Cremer, X.~Li, and D.~Duvenaud.
\newblock Inference suboptimality in variational autoencoders.
\newblock In \emph{International Conference on Machine Learning}, pages
  1078--1086. PMLR, 2018.

\bibitem[Dayan et~al.(1995)Dayan, Hinton, Neal, and Zemel]{dayan1995helmholtz}
P.~Dayan, G.~E. Hinton, R.~M. Neal, and R.~S. Zemel.
\newblock The helmholtz machine.
\newblock \emph{Neural computation}, 7\penalty0 (5):\penalty0 889--904, 1995.

\bibitem[Deng et~al.(2009)Deng, Dong, Socher, Li, Li, and
  Fei-Fei]{deng2009imagenet}
J.~Deng, W.~Dong, R.~Socher, L.-J. Li, K.~Li, and L.~Fei-Fei.
\newblock Imagenet: A large-scale hierarchical image database.
\newblock In \emph{2009 IEEE conference on computer vision and pattern
  recognition}, pages 248--255. Ieee, 2009.

\bibitem[Flamich et~al.(2020)Flamich, Havasi, and
  Hern{\'a}ndez-Lobato]{flamich2020compressing}
G.~Flamich, M.~Havasi, and J.~M. Hern{\'a}ndez-Lobato.
\newblock Compressing images by encoding their latent representations with
  relative entropy coding.
\newblock \emph{Advances in Neural Information Processing Systems},
  33:\penalty0 16131--16141, 2020.

\bibitem[Gao et~al.(2021)Gao, You, Pan, Han, Zhang, Dai, and
  Lee]{gao2021neural}
G.~Gao, P.~You, R.~Pan, S.~Han, Y.~Zhang, Y.~Dai, and H.~Lee.
\newblock Neural image compression via attentional multi-scale back projection
  and frequency decomposition.
\newblock In \emph{Proceedings of the IEEE/CVF International Conference on
  Computer Vision}, pages 14677--14686, 2021.

\bibitem[Guo et~al.(2021{\natexlab{a}})Guo, Zhang, Feng, and
  Chen]{guo2021causal}
Z.~Guo, Z.~Zhang, R.~Feng, and Z.~Chen.
\newblock Causal contextual prediction for learned image compression.
\newblock \emph{IEEE Transactions on Circuits and Systems for Video
  Technology}, 2021{\natexlab{a}}.

\bibitem[Guo et~al.(2021{\natexlab{b}})Guo, Zhang, Feng, and Chen]{guo2021soft}
Z.~Guo, Z.~Zhang, R.~Feng, and Z.~Chen.
\newblock Soft then hard: Rethinking the quantization in neural image
  compression.
\newblock In \emph{International Conference on Machine Learning}, pages
  3920--3929. PMLR, 2021{\natexlab{b}}.

\bibitem[He et~al.(2021)He, Zheng, Sun, Wang, and Qin]{he2021checkerboard}
D.~He, Y.~Zheng, B.~Sun, Y.~Wang, and H.~Qin.
\newblock Checkerboard context model for efficient learned image compression.
\newblock In \emph{Proceedings of the IEEE/CVF Conference on Computer Vision
  and Pattern Recognition}, pages 14771--14780, 2021.

\bibitem[He et~al.(2022)He, Yang, Peng, Ma, Qin, and Wang]{he2022elic}
D.~He, Z.~Yang, W.~Peng, R.~Ma, H.~Qin, and Y.~Wang.
\newblock Elic: Efficient learned image compression with unevenly grouped
  space-channel contextual adaptive coding.
\newblock \emph{arXiv preprint arXiv:2203.10886}, 2022.

\bibitem[Hinton and Van~Camp(1993)]{hinton1993keeping}
G.~E. Hinton and D.~Van~Camp.
\newblock Keeping the neural networks simple by minimizing the description
  length of the weights.
\newblock In \emph{Proceedings of the sixth annual conference on Computational
  learning theory}, pages 5--13, 1993.

\bibitem[Hinton et~al.(1995)Hinton, Dayan, Frey, and Neal]{hinton1995wake}
G.~E. Hinton, P.~Dayan, B.~J. Frey, and R.~M. Neal.
\newblock The" wake-sleep" algorithm for unsupervised neural networks.
\newblock \emph{Science}, 268\penalty0 (5214):\penalty0 1158--1161, 1995.

\bibitem[Hu et~al.(2021)Hu, Lu, and Xu]{hu2021fvc}
Z.~Hu, G.~Lu, and D.~Xu.
\newblock Fvc: A new framework towards deep video compression in feature space.
\newblock In \emph{Proceedings of the IEEE/CVF Conference on Computer Vision
  and Pattern Recognition}, pages 1502--1511, 2021.

\bibitem[Kingma and Welling(2013)]{kingma2013auto}
D.~P. Kingma and M.~Welling.
\newblock Auto-encoding variational bayes.
\newblock \emph{arXiv preprint arXiv:1312.6114}, 2013.

\bibitem[Kodak(1993)]{kodak}
E.~Kodak.
\newblock Kodak lossless true color image suite.
\newblock \url{http://r0k.us/graphics/kodak/}, 1993.

\bibitem[Li et~al.(2021)Li, Li, and Lu]{li2021deep}
J.~Li, B.~Li, and Y.~Lu.
\newblock Deep contextual video compression.
\newblock \emph{Advances in Neural Information Processing Systems}, 34, 2021.

\bibitem[Loshchilov and Hutter(2016)]{loshchilov2016sgdr}
I.~Loshchilov and F.~Hutter.
\newblock Sgdr: Stochastic gradient descent with warm restarts.
\newblock \emph{arXiv preprint arXiv:1608.03983}, 2016.

\bibitem[Lu et~al.(2019)Lu, Ouyang, Xu, Zhang, Cai, and Gao]{lu2019dvc}
G.~Lu, W.~Ouyang, D.~Xu, X.~Zhang, C.~Cai, and Z.~Gao.
\newblock Dvc: An end-to-end deep video compression framework.
\newblock In \emph{Proceedings of the IEEE/CVF Conference on Computer Vision
  and Pattern Recognition}, pages 11006--11015, 2019.

\bibitem[Lu et~al.(2020)Lu, Zhang, Ouyang, Chen, Gao, and Xu]{lu2020end}
G.~Lu, X.~Zhang, W.~Ouyang, L.~Chen, Z.~Gao, and D.~Xu.
\newblock An end-to-end learning framework for video compression.
\newblock \emph{IEEE transactions on pattern analysis and machine
  intelligence}, 43\penalty0 (10):\penalty0 3292--3308, 2020.

\bibitem[Minnen and Singh(2020)]{minnen2020channel}
D.~Minnen and S.~Singh.
\newblock Channel-wise autoregressive entropy models for learned image
  compression.
\newblock In \emph{2020 IEEE International Conference on Image Processing
  (ICIP)}, pages 3339--3343. IEEE, 2020.

\bibitem[Minnen et~al.(2018)Minnen, Ball{\'e}, and Toderici]{minnen2018joint}
D.~Minnen, J.~Ball{\'e}, and G.~D. Toderici.
\newblock Joint autoregressive and hierarchical priors for learned image
  compression.
\newblock \emph{Advances in neural information processing systems}, 31, 2018.

\bibitem[Mnih and Gregor(2014)]{mnih2014neural}
A.~Mnih and K.~Gregor.
\newblock Neural variational inference and learning in belief networks.
\newblock In \emph{International Conference on Machine Learning}, pages
  1791--1799. PMLR, 2014.

\bibitem[Mnih and Rezende(2016)]{mnih2016variational}
A.~Mnih and D.~Rezende.
\newblock Variational inference for monte carlo objectives.
\newblock In \emph{International Conference on Machine Learning}, pages
  2188--2196. PMLR, 2016.

\bibitem[Mohamed et~al.(2020)Mohamed, Rosca, Figurnov, and
  Mnih]{mohamed2020monte}
S.~Mohamed, M.~Rosca, M.~Figurnov, and A.~Mnih.
\newblock Monte carlo gradient estimation in machine learning.
\newblock \emph{J. Mach. Learn. Res.}, 21\penalty0 (132):\penalty0 1--62, 2020.

\bibitem[Nowozin(2018)]{nowozin2018debiasing}
S.~Nowozin.
\newblock Debiasing evidence approximations: On importance-weighted
  autoencoders and jackknife variational inference.
\newblock In \emph{International conference on learning representations}, 2018.

\bibitem[Paulus et~al.(2020)Paulus, Maddison, and Krause]{paulus2020rao}
M.~B. Paulus, C.~J. Maddison, and A.~Krause.
\newblock Rao-blackwellizing the straight-through gumbel-softmax gradient
  estimator.
\newblock \emph{arXiv preprint arXiv:2010.04838}, 2020.

\bibitem[Rainforth et~al.(2018)Rainforth, Kosiorek, Le, Maddison, Igl, Wood,
  and Teh]{rainforth2018tighter}
T.~Rainforth, A.~Kosiorek, T.~A. Le, C.~Maddison, M.~Igl, F.~Wood, and Y.~W.
  Teh.
\newblock Tighter variational bounds are not necessarily better.
\newblock In \emph{International Conference on Machine Learning}, pages
  4277--4285. PMLR, 2018.

\bibitem[Roeder et~al.(2017)Roeder, Wu, and Duvenaud]{roeder2017sticking}
G.~Roeder, Y.~Wu, and D.~K. Duvenaud.
\newblock Sticking the landing: Simple, lower-variance gradient estimators for
  variational inference.
\newblock \emph{Advances in Neural Information Processing Systems}, 30, 2017.

\bibitem[Ryder et~al.(2022)Ryder, Zhang, Kang, and Zhang]{ryder2022split}
T.~Ryder, C.~Zhang, N.~Kang, and S.~Zhang.
\newblock Split hierarchical variational compression.
\newblock In \emph{Proceedings of the IEEE/CVF Conference on Computer Vision
  and Pattern Recognition}, pages 386--395, 2022.

\bibitem[Serban et~al.(2017)Serban, Sordoni, Lowe, Charlin, Pineau, Courville,
  and Bengio]{serban2017hierarchical}
I.~Serban, A.~Sordoni, R.~Lowe, L.~Charlin, J.~Pineau, A.~Courville, and
  Y.~Bengio.
\newblock A hierarchical latent variable encoder-decoder model for generating
  dialogues.
\newblock In \emph{Proceedings of the AAAI Conference on Artificial
  Intelligence}, volume~31, 2017.

\bibitem[Song et~al.(2020)Song, Xu, Yu, Zhou, Shao, and Yu]{song2020infomax}
Y.~Song, M.~Xu, L.~Yu, H.~Zhou, S.~Shao, and Y.~Yu.
\newblock Infomax neural joint source-channel coding via adversarial bit flip.
\newblock In \emph{Proceedings of the AAAI Conference on Artificial
  Intelligence}, volume~34, pages 5834--5841, 2020.

\bibitem[Theis and Agustsson(2021)]{theis2021advantages}
L.~Theis and E.~Agustsson.
\newblock On the advantages of stochastic encoders.
\newblock \emph{arXiv preprint arXiv:2102.09270}, 2021.

\bibitem[Theis and Ho(2021)]{theis2021importance}
L.~Theis and J.~Ho.
\newblock Importance weighted compression.
\newblock In \emph{Neural Compression: From Information Theory to
  Applications--Workshop@ ICLR 2021}, 2021.

\bibitem[Theis et~al.(2017)Theis, Shi, Cunningham, and
  Husz{\'a}r]{theis2017lossy}
L.~Theis, W.~Shi, A.~Cunningham, and F.~Husz{\'a}r.
\newblock Lossy image compression with compressive autoencoders.
\newblock \emph{arXiv preprint arXiv:1703.00395}, 2017.

\bibitem[Toderici et~al.(2015)Toderici, O'Malley, Hwang, Vincent, Minnen,
  Baluja, Covell, and Sukthankar]{toderici2015variable}
G.~Toderici, S.~M. O'Malley, S.~J. Hwang, D.~Vincent, D.~Minnen, S.~Baluja,
  M.~Covell, and R.~Sukthankar.
\newblock Variable rate image compression with recurrent neural networks.
\newblock \emph{arXiv preprint arXiv:1511.06085}, 2015.

\bibitem[Toderici et~al.(2017)Toderici, Vincent, Johnston, Jin~Hwang, Minnen,
  Shor, and Covell]{toderici2017full}
G.~Toderici, D.~Vincent, N.~Johnston, S.~Jin~Hwang, D.~Minnen, J.~Shor, and
  M.~Covell.
\newblock Full resolution image compression with recurrent neural networks.
\newblock In \emph{Proceedings of the IEEE conference on Computer Vision and
  Pattern Recognition}, pages 5306--5314, 2017.

\bibitem[Townsend et~al.(2018)Townsend, Bird, and
  Barber]{townsend2018practical}
J.~Townsend, T.~Bird, and D.~Barber.
\newblock Practical lossless compression with latent variables using bits back
  coding.
\newblock In \emph{International Conference on Learning Representations}, 2018.

\bibitem[Tucker et~al.(2018)Tucker, Lawson, Gu, and Maddison]{tucker2018doubly}
G.~Tucker, D.~Lawson, S.~Gu, and C.~J. Maddison.
\newblock Doubly reparameterized gradient estimators for monte carlo
  objectives.
\newblock 2018.

\bibitem[Van Den~Oord et~al.(2017)Van Den~Oord, Vinyals, et~al.]{van2017neural}
A.~Van Den~Oord, O.~Vinyals, et~al.
\newblock Neural discrete representation learning.
\newblock \emph{Advances in neural information processing systems}, 30, 2017.

\bibitem[Williams(1992)]{williams1992simple}
R.~J. Williams.
\newblock Simple statistical gradient-following algorithms for connectionist
  reinforcement learning.
\newblock \emph{Machine learning}, 8\penalty0 (3):\penalty0 229--256, 1992.

\bibitem[Xie et~al.(2021)Xie, Cheng, and Chen]{xie2021enhanced}
Y.~Xie, K.~L. Cheng, and Q.~Chen.
\newblock Enhanced invertible encoding for learned image compression.
\newblock In \emph{Proceedings of the 29th ACM International Conference on
  Multimedia}, pages 162--170, 2021.

\bibitem[Yang et~al.(2020)Yang, Bamler, and Mandt]{yang2020improving}
Y.~Yang, R.~Bamler, and S.~Mandt.
\newblock Improving inference for neural image compression.
\newblock \emph{Advances in Neural Information Processing Systems},
  33:\penalty0 573--584, 2020.

\bibitem[Zhu et~al.(2022)Zhu, Song, Gao, Zheng, and Shen]{zhu2022unified}
X.~Zhu, J.~Song, L.~Gao, F.~Zheng, and H.~T. Shen.
\newblock Unified multivariate gaussian mixture for efficient neural image
  compression.
\newblock \emph{arXiv preprint arXiv:2203.10897}, 2022.

\bibitem[Zhu et~al.(2021)Zhu, Yang, and Cohen]{zhu2021transformer}
Y.~Zhu, Y.~Yang, and T.~Cohen.
\newblock Transformer-based transform coding.
\newblock In \emph{International Conference on Learning Representations}, 2021.

\end{thebibliography}

\newpage

\appendix

\section{Appendix}

\subsection{ELBO and Bits-Back Coding}
\label{sec:bb}
It is well known that the ELBO $\mathcal{L}$ is the minus overall bitrate for bits-back coding in compression \citep{hinton1993keeping, hinton1995wake, chen2017variational}, and the entropy of variational posterior is exactly the bits-back rate itself. For this reason, earlier works \citep{townsend2018practical, yang2020improving} point out that \citep{balle2018variational, minnen2018joint} waste bits for not using bits-back coding on $\bm{z}$. However, during training the $\mathbb{E}_{q(\bmt{z}|\bmt{y})}[\log q(\bmt{z}|\bmt{y})]$ is constant. And this means that this term does not have impact on the optimization procedure. And due to the deterministic inference, the $\log q(\bar{\bm{z}}|\bar{\bm{y}})$ is $0$, which means that the bitrate saved by bits-back coding is $0$. In this sense, \citep{balle2018variational, minnen2018joint} is also optimal in bits-back coding perspective, although no actual bits-back coding is performed. In fact, there is no space for bits-back coding so long as encoder is deterministic. Since we can view deterministic encoder as a posterior distribution with mass $1$ on a single point. And then the posterior's entropy is always $0$.

\subsection{Plate Notations of Generative Models}
\label{sec:plate2}

\begin{figure}[thb]
\begin{center}
    \includegraphics[width=0.5\linewidth]{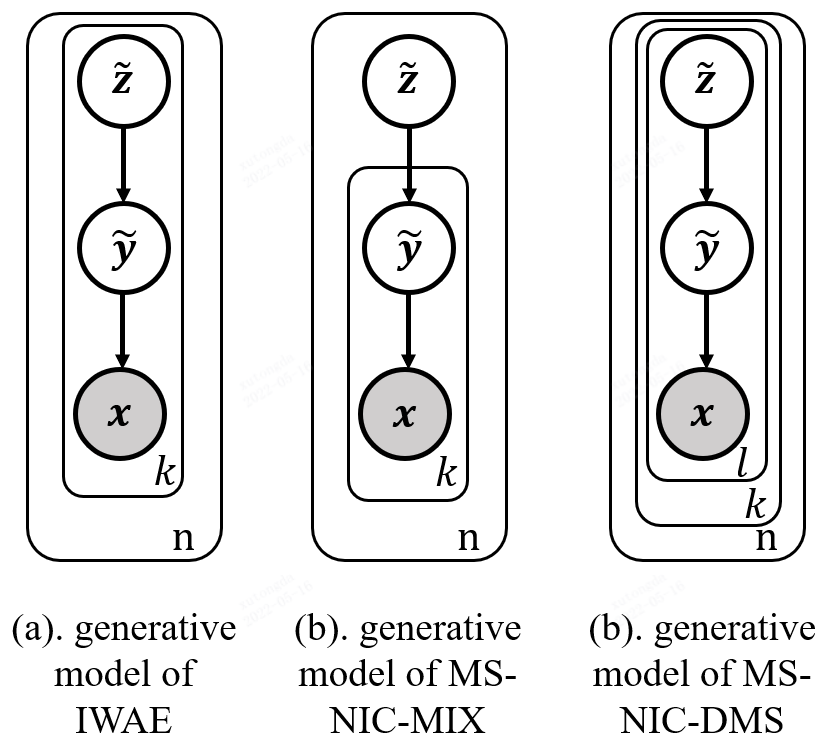}
    \label{fig:plate2}
    \caption{The generative model of Fig.~\ref{fig:plate} (d)-(f). The generative models show how we compute data likelihood for multiple-sample approaches, not how the image is actually generated in nature. For clarity, we omit the parameters.}
\end{center}
\end{figure}

Fig.~\ref{fig:plate2} shows the generative models of Fig.~\ref{fig:plate} (d)-(f). Note that we have only one unique sample of $\bm{x}$ inside the $n$ plate, but it is repeated $k$ times for IWAE, MS-NIC-MIX and $k\times l$ times for MS-NIC-DMS. Similarly, k samples of $\bmt{y}$ is repeated $l$ times and $l$ samples of $\bmt{z}$ is repeated $k$ times for MS-NIC-DMS.

\subsection{Proof on the Properties of MS-NIC-MIX and MS-NIC-DMS}
\label{app:pf}
In this section, we add the $q(\bmt{z}|\bmt{y}), q(\bmt{y}|\bm{x})$ back to equations for clarity of the proof. This makes the notations slightly different from Eq.~\ref{eq:yelbort} and Eq.~\ref{eq:yzelbo18hat}. Note that divide by $q(\bmt{z}|\bmt{y}), q(\bmt{y}|\bm{x})$ does not effect the value of equation, and add $\log q(\bmt{z}|\bmt{y}), \log q(\bmt{y}|\bm{x})$ does not effect the value of equation.

For MS-NIC-MIX to be a reasonably better approach to apply over \citep{balle2018variational}, we show that $\mathcal{L}^{MIX}_k$ satisfies following properties:
\begin{enumerate}
    \item $\log p(x) \ge \mathcal{L}^{MIX}_k $
    \item $\mathcal{L}^{MIX}_k\ge  \mathcal{L}^{MIX}_m$ for $k \ge m$
\end{enumerate}

We can show 1. $\log p(x) \ge \mathcal{L}^{MIX}_k $ by applying Jensen's inequality twice:

\begin{equation}
  \begin{array}{l}
  \mathcal{L}_{k}^{MIX} = \mathbb{E}_{q_{\phi}(\bmt{z}|\bm{x})}[\mathbb{E}_{q_{\phi}(\bmt{y}_{1:k}|\bm{x})}[\log \frac{1}{k}\overset{k}{\underset{i}{\sum}} \frac{p(\bm{x}|\bmt{y}_{i})p(\bmt{y}_{i}|\bmt{z})}{q(\bmt{y}_i|\bm{x})}|\bmt{z}] + \log p(\bmt{z}) - \log q(\bmt{z}|\bm{x})] \\ 
  \hspace{3em} \le \mathbb{E}_{q_{\phi}(\bmt{z}|\bm{x})}[\log (\frac{1}{k}\overset{k}{\underset{i}{\sum}}\mathbb{E}_{q_{\phi}(\bmt{y}_{1:k}|\bm{x})}[  \frac{p(\bm{x}|\bmt{y}_{i})p(\bmt{y}_{i}|\bmt{z})}{q(\bmt{y}_i|\bm{x})}|\bmt{z}]) + \log p(\bmt{z}) - \log q(\bmt{z}|\bm{x})] \\
  \hspace{3em} = \mathbb{E}_{q_{\phi}(\bmt{z}|\bm{x})}[\log p(\bm{x}|\bmt{z}) + \log p(\bmt{z}) - \log q(\bmt{z}|\bm{x})] \\
  \hspace{3em} \le \log (\mathbb{E}_{q_{\phi}(\bmt{z}|\bm{x})}[\frac{p(\bm{x}|\bmt{z})p(\bmt{z})}{q(\bmt{z}|\bm{x})}]) \\
  \hspace{3em} = \log p(x)
  \end{array}
\label{eq:mixproof1}
\end{equation}

We can show 2. $\mathcal{L}^{MIX}_k\ge  \mathcal{L}^{MIX}_m$ for $k \ge m$ by borrowing the Theorem 1 from IWAE paper:

\begin{equation}
  \begin{array}{l}
    k \ge m \Rightarrow \mathbb{E}_{q(\bm{h}_i|\bm{x})}[\log \frac{1}{k}\overset{k}{\underset{i}{\sum}}\frac{p(\bm{h}_i | \bm{x})p(\bm{h}_i)}{q(\bm{h}_i|\bm{x})}] \ge \mathbb{E}_{q(\bm{h}_i|\bm{x})}[\log \frac{1}{m}\overset{m}{\underset{i}{\sum}}\frac{p(\bm{h}_i | \bm{x})p(\bm{h}_i)}{q(\bm{h}_i|\bm{x})}] 
    \end{array}
\label{eq:mixproof2}
\end{equation}

Applying Eq.~\ref{eq:mixproof2} to the internal part of $\mathcal{L}^{MIX}_k$, when $k \ge m$, we have:

\begin{equation}
  \begin{array}{l}
  \mathcal{L}_{k}^{MIX} = \mathbb{E}_{q_{\phi}(\bmt{z}|\bm{x})}[\mathbb{E}_{q_{\phi}(\bmt{y}_{1:k}|\bm{x})}[\log \frac{1}{k}\overset{k}{\underset{i}{\sum}} \frac{p(\bm{x}|\bmt{y}_{i})p(\bmt{y}_{i}|\bmt{z})}{q(\bmt{y}_i|\bm{x})}|\bmt{z}] + \log p(\bmt{z}) - \log q(\bmt{z}|\bm{x})] \\ 
  \hspace{3em} \ge \mathbb{E}_{q_{\phi}(\bmt{z}|\bm{x})}[\mathbb{E}_{q_{\phi}(\bmt{y}_{1:m}|\bm{x})}[\log \frac{1}{m}\overset{m}{\underset{i}{\sum}} \frac{p(\bm{x}|\bmt{y}_{i})p(\bmt{y}_{i}|\bmt{z})}{q(\bmt{y}_i|\bm{x})}|\bmt{z}] + \log p(\bmt{z}) - \log q(\bmt{z}|\bm{x})] \\ 
  \hspace{3em} = \mathcal{L}^{MIX}_m

  \end{array}
\label{eq:mixproof3}
\end{equation}

For MS-NIC-DMS to be a reasonably better approach to apply over \citep{balle2018variational} and MS-NIC-MIX, we show that $\mathcal{L}^{DMS}_{k, l}$ statisfies following properties:

\begin{enumerate}
    \item $\log p(x) \ge \mathcal{L}^{DMS}_{k, l} $
    \item $\mathcal{L}^{DMS}_{k, l}\ge  \mathcal{L}^{DMS}_{m, n}$ for $k \ge m, l \ge n$
    \item $\mathcal{L}^{DMS}_{k, l} \ge \mathcal{L}^{MIX}_{k}$
    \item $\mathcal{L}^{DMS}_{k, l} \rightarrow \log p(x)$ as $k, l \rightarrow \infty$, under the assumption that $\log \frac{p(\bm{x}|\bmt{y}_{i})p(\bmt{y}_{i}|\bmt{z}_{j})}{q(\bmt{y}_i|\bm{x})}$ and $\log \frac{p(\bm{x}|\bmt{z}_{j})p(\bmt{z}_{j})}{q(\bmt{z}_j|\bm{x})}$ are bounded.
\end{enumerate}

Similar to MS-NIC-DMS, we can show 1.$\log p(x) \ge \mathcal{L}^{DMS}_{k, l} $ by applying Jensen's inequality twice: 

\begin{equation}
  \begin{array}{l}
  \mathcal{L}_{k,l}^{DMS} = \mathbb{E}_{q_{\phi}(\bmt{z}_{1:l}|\bm{x})}[\log \frac{1}{l}\overset{l}{\underset{j}{\sum}} \exp{(\mathbb{E}_{q_{\phi}(\bmt{y}_{1:k}|\bm{x})}[\log \frac{1}{k}\overset{k}{\underset{i}{\sum}} \frac{p(\bm{x}|\bmt{y}_{i})p(\bmt{y}_{i}|\bmt{z}_{j})}{q(\bmt{y}_i|\bm{x})}|\bmt{z}_{j}])}p(\bmt{z}_{j})/q(\bmt{z}_j|\bm{x})] \\
  \hspace{3em} \le 
  \mathbb{E}_{q_{\phi}(\bmt{z}_{1:l}|\bm{x})}[\log \frac{1}{l}\overset{l}{\underset{j}{\sum}} \exp{\log (\frac{1}{k}\overset{k}{\underset{i}{\sum}}\mathbb{E}_{q_{\phi}(\bmt{y}_{1:k}|\bm{x})}[ p(\bm{x}|\bmt{y}_{i})p(\bmt{y}_{i}|\bmt{z}_{j})|\bmt{z}_{j}])}p(\bmt{z}_{j})/q(\bmt{z}_j|\bm{x})] \\
  \hspace{3em} =
  \mathbb{E}_{q_{\phi}(\bmt{z}_{1:l}|\bm{x})}[\log \frac{1}{l}\overset{l}{\underset{j}{\sum}} \frac{p(\bm{x}|\bmt{z}_{j})p(\bmt{z}_{j})}{q(\bmt{z}_j|\bm{x})}] \\
  \hspace{3em} \le
  \log \frac{1}{l}\overset{l}{\underset{j}{\sum}} \mathbb{E}_{q_{\phi}(\bmt{z}_{1:l}|\bm{x})}[  \frac{p(\bm{x}|\bmt{z}_{j})p(\bmt{z}_{j})}{q(\bmt{z}_j|\bm{x})}] \\
  \hspace{3em} = \log p(\bm{x})
  \end{array}
\label{eq:dmsproof1}
\end{equation}

Also similar to MS-NIC-MIX, we can borrow conclusion from IWAE (Eq.~\ref{eq:mixproof2}) and apply it twice to show 2. $\mathcal{L}^{DMS}_{k, l}\ge  \mathcal{L}^{DMS}_{m, n}$ for $k \ge m, l \ge n$:

\begin{equation}
  \begin{array}{l}
  \mathcal{L}_{k,l}^{DMS} = \mathbb{E}_{q_{\phi}(\bmt{z}_{1:l}|\bm{x})}[\log \frac{1}{l}\overset{l}{\underset{j}{\sum}} \exp{(\mathbb{E}_{q_{\phi}(\bmt{y}_{1:k}|\bm{x})}[\log \frac{1}{k}\overset{k}{\underset{i}{\sum}} \frac{p(\bm{x}|\bmt{y}_{i})p(\bmt{y}_{i}|\bmt{z}_{j})}{q(\bmt{y}_i|\bm{x})}|\bmt{z}_{j}])}p(\bmt{z}_{j})/q(\bmt{z}_j|\bm{x})] \\
  \hspace{3em} \ge
 \mathbb{E}_{q_{\phi}(\bmt{z}_{1:l}|\bm{x})}[\log \frac{1}{l}\overset{l}{\underset{j}{\sum}} \exp{(\mathbb{E}_{q_{\phi}(\bmt{y}_{1:m}|\bm{x})}[\log \frac{1}{m}\overset{m}{\underset{i}{\sum}} \frac{p(\bm{x}|\bmt{y}_{i})p(\bmt{y}_{i}|\bmt{z}_{j})}{q(\bmt{y}_i|\bm{x})}|\bmt{z}_{j}])}p(\bmt{z}_{j})/q(\bmt{z}_j|\bm{x})] \\
  \hspace{3em} \ge
 \mathbb{E}_{q_{\phi}(\bmt{z}_{1:n}|\bm{x})}[\log \frac{1}{n}\overset{n}{\underset{j}{\sum}} \exp{(\mathbb{E}_{q_{\phi}(\bmt{y}_{1:m}|\bm{x})}[\log \frac{1}{m}\overset{m}{\underset{i}{\sum}} \frac{p(\bm{x}|\bmt{y}_{i})p(\bmt{y}_{i}|\bmt{z}_{j})}{q(\bmt{y}_i|\bm{x})}|\bmt{z}_{j}])}p(\bmt{z}_{j})/q(\bmt{z}_j|\bm{x})] \\
   \hspace{3em} = \mathcal{L}^{DMS}_{m, n}
  \end{array}
\label{eq:dmsproof2}
\end{equation}

With 2. $\mathcal{L}^{DMS}_{k, l}\ge  \mathcal{L}^{DMS}_{m, n}$ for $k \ge m, l \ge n$ holds, we can show 3. $\mathcal{L}^{DMS}_{k, l} \ge \mathcal{L}^{MIX}_{k}$ immediately as $\mathcal{L}^{DMS}_{k, l} \ge \mathcal{L}^{DMS}_{k, 1} = \mathcal{L}^{MIX}_{k}$.

To show 4. $\mathcal{L}^{DMS}_{k, l} \rightarrow \log p(x)$ as $k, l \rightarrow \infty$, we first define intermediate variables $W_k, \tilde{M}_{k, l}, M_{k, l}$:

\begin{equation}
  \begin{array}{l}
    \hspace{0.5em} W_{k} = \frac{1}{k}\overset{k}{\underset{i}{\sum}} \frac{p(\bm{x}|\bmt{y}_{i})p(\bmt{y}_{i}|\bmt{z}_{j})}{q(\bmt{y}_i|\bm{x})} \\
    \tilde{M}_{k, l} =  \frac{1}{l}\overset{l}{\underset{j}{\sum}}\frac{p(\bm{x}|\bmt{z}_{j})p(\bmt{z}_{j})}{q(\bmt{z}_j|\bm{x})} \\
    M_{k, l} =  \frac{1}{l}\overset{l}{\underset{j}{\sum}} \exp{(\mathbb{E}_{q_{\phi}(\bmt{y}_{1:k}|\bm{x})}[\log W_{k}|\bmt{z}_{j}])}p(\bmt{z}_{j})/q(\bmt{z}_j|\bm{x})
  \end{array}
\label{eq:dmsproof3}
\end{equation}

Under the assumption that $\log p(\bm{x}|\bmt{y}_{i})p(\bmt{y}_{i}|\bmt{z}_{j})/q(\bmt{y}_i|\bm{x})$ is bounded, from the strong law of large number, we have $W_{k} \xrightarrow{a.s.}  p(\bm{x}|\bmt{z}_j)$ (Eq.~\ref{eq:dmsproof4}). Then we have $\mathbb{E}[\log W_k|\bmt{z}_j] \rightarrow \log p(\bm{x}|\bmt{z}_j)$.
\begin{equation}
  \begin{array}{l}
  W_{k} \xrightarrow{a.s.} \mathbb{E}_{q(\bmt{y}_i|\bm{x})}[\frac{p(\bm{x}|\bmt{y}_{i})p(\bmt{y}_{i}|\bmt{z}_{j})}{q(\bmt{y}_i|\bm{x})}|\bmt{z}_{j}] = \int q(\bmt{y}_i|\bm{x}) \frac{p(\bm{x}|\bmt{y}_{i})p(\bmt{y}_{i}|\bmt{z}_{j})}{q(\bmt{y}_i|\bm{x})} d\bmt{y}_i = p(\bm{x}|\bmt{z}_j)
  \end{array}
\label{eq:dmsproof4}
\end{equation}

Moreover, as $\mathbb{E}[\log W_k|\bmt{z}_j] \rightarrow \log p(\bm{x}|\bmt{z}_j)$, we have $M_{k, l} \rightarrow \tilde{M}_{k, l}$. This means that $\forall \epsilon > 0, \exists k,l, s.t. |M_{k, l} - \tilde{M}_{k, l}| < \epsilon$. And thus we have $|\mathbb{E}[M_{k, l}] - \mathbb{E}[\tilde{M}_{k, l}]| \le \mathbb{E}[|M_{k, l} - \tilde{M}_{k, l}|] < \epsilon \rightarrow 0$. Then we have $|E[M_{k, l}] - p(\bm{x})| \le |E[M_{k, l}] - E[\tilde{M}_{k, l}]| + |E[\tilde{M}_{k, l}] - p(\bm{x})| \rightarrow 0$, and thus $E[M_{k, l}] \rightarrow p(\bm{x})$. Finally we have $\mathbb{E}[\log M_{k, l}] = L^{DMS}_{k, l} \rightarrow \log p(\bm{x})$.

\subsection{Effects of Sample Size}
\label{ap:samplesize}

When comparing the R-D performance of models trained with a single $\lambda$, we use R-D cost as our metric. The R-D cost is simply computed as bpp $ + \lambda$ MSE, where bpp is a short of bits-per-pixel, and MSE is a short of mean square error. The lower the R-D cost is, the better the R-D performance is. Another way to interpret R-D cost is to view it as the ELBO with constant offset. Then the $\lambda$ MSE is connected to the log likelihood of a Gaussian distribution whose mean is the output of decoder and sigma is determined by $\lambda$. Note that R-D cost is only comparable when $\lambda$ is the same. 

Tab.~\ref{tab:ablation} shows the effect of sample size to MS-NIC. Moreover, we compare the na\"ive increase of batch size versus multiple importance weighted samples. As shown by the table, increasing the batch size $\times 3 - 16$ only slightly affects the R-D cost (from $1.017$ to $1.013$). However, the MS-NIC-MIX can achieve R-D cost of $0.9988$ with sample size 8, and MS-NIC-DMS can achieve $0.9954$ with sample size 16. This means that MS-NIC is effective over the baseline and vanilla batch size increases. It is also noteworthy that we have not observed inference model training failure as sample size increase. While MS-NIC also suffers from gradient SNR vanishing problem, a sample size of $16$ is probably not large enough to make it evident. Limited by computational power, we can not raise sample size by several magnitudes as \citep{rainforth2018tighter} does with small model.

\begin{table}[ht]
\centering
\caption{Effect of sample size in MS-NIC.}
\vspace{2mm}
\label{tab:ablation}
\begin{tabular}{lccccc}
\toprule
                      & Sample/Batch Size & bpp & MSE & PSNR (db) & R-D cost          \\ \midrule
Baseline \citep{balle2018variational} & - & 0.5273 & 32.61 & 33.28 & 1.017 \\ \midrule
Baseline-BigBatch      & $\times$3 & 0.5308 & 32.51 & 33.31	& 1.018 \\
                & $\times$5 & 0.5285 & 32.51 & 33.30 & 1.016 \\
                & $\times$8 & 0.5279 & 32.37 & 33.34 & 1.013 \\
                & $\times$16 & 0.5321 & 32.12 &	33.38 & 1.014 \\ \midrule
IWAE \citep{burda2016importance}    & 3 & 0.9128 & 32.46 & 33.28 & 1.400 \\
                & 5 & 0.7903 & 31.73 & 33.40 & 1.266 \\
                & 8 & 0.9477 & 31.48 & 33.44 & 1.420 \\
                & 16 & 1.273 & 31.69 & 33.40 & 1.748 \\ \midrule
MS-NIC-MIX      & 3 & 0.5238 & 31.80 & 33.40 & 1.000 \\
                & 5 & 0.5259 & 31.84 & 33.38 & 1.003 \\
                & 8 & 0.5260 & 31.52 & 33.44 & 0.9988 \\
                & 16 & 0.5256 & 32.48 & 33.29 & 1.013 \\ \midrule
MS-NIC-DMS       & 3, 3 & 0.5247 & 32.39 & 33.30 & 1.010 \\
                & 5, 5 & 0.5230 & 31.84 & 33.39 & 1.001 \\
                & 8, 8 & 0.5255	& 31.55 & 33.43 & 0.9989 \\
                & 16, 16 & 0.5249 &	31.38 &	33.46 & 0.9954 \\ \bottomrule
\end{tabular}
\end{table}

\subsection{Detailed Experimental Settings}
\label{ap:detailsset}
All the experiments are conducted on a computer with Intel(R) Xeon(R) CPU E5-2620 v4 @ 2.10GHz and $8\times$ Nvidia(R) TitanXp. All the training scripts are implemented with Pytorch 1.7 and CUDA 9.0. For experiments with single-sample, we adopt Adam optimizer with $\beta_1=0.90, \beta_2=0.95, lr=1e^{-4}$. For experiments with multiple-sample/big batch, we scale $lr$ linearly with sample size. All the models are trained for $2000$ epochs with the settings in Sec.~5.1. For first $200$ epochs, we adopt cosine annealing \citep{loshchilov2016sgdr} to schedule learning rate. It takes around $1-2$ days to train models based on \citet{balle2018variational}, and $3-5$ days to train models on \citet{cheng2020learned}. Note that our multiple-sample approaches' training time does not scale linearly with sample size, as we perform sampling on posterior, and the variational encoder only computes parameter of posterior parameters once. Further, we provide the pytorch style sudo code for implementation guidance of MS-NIC-MIX and MS-NIC-DMS.
\begin{small}
\begin{lstlisting}
import torch
from torch.nn import functional as F

def IWAELoss(minus_elbo):
    '''
    args
    ----
    minus_elbo: tensor, [b, k], which is R + \lambda D

    return
    ------
    local iwae loss
    '''
    # this is the minus ELBO related to y part, 
    # to get the real ELBO:
    log_weights = - minus_elbo.detach()
    # no gradient given to weights
    weights = F.softmax(log_weights, dim=1) # B, K
    loss_b = torch.sum(minus_elbo * weights, dim=1, keepdim=False)
    loss_iwae = torch.mean(loss_b)
    return loss_iwae

def DMSLoss(x, x_hat, y_likelihood, z_likelihood, lam):
    '''
    args
    ----
    x: original image: [b, c, h, w]
    x_hat: reconstructed image: [b, k, c, h, w], k is the
        number of samples
    y_likelihood: [b, 192/320, h//8, w//8, k^2],
        as original paper of [Balle et al. 2018], the number of 
        channels 192/320 is determined by lambda, k^2 is the
        number of samples in DMS setting, with MS-NIC-MIX,
        this k^2 is k
    z_likelihood: [b, 128/192, h//64, w//64, k], as original
        paper of [Balle et al. 2018], the number of channels
        128/192 is determined by lambda,
        k is the number of samples

    return
    ------
    total iwae loss
    '''
    b, c, h, w = x.shape
    k = x_hat.shape[0] // x.shape[0]
    x = torch.repeat_interleave(x, repeats=k, dim=0)
    x = x.reshape(b, k, c, h, w)
    x_hat = x_hat.reshape(b, k, c, h, w)
    d_loss = torch.mean(lam * 65025 * (x - x_hat)**2, dim=(2,3,4),
                        keepdim=False)
    yz_loss = -torch.sum(torch.log2(y_likelihood), dim=(1,2,3)).\
              reshape(b, -1) / (h * w)
    z_loss = -torch.sum(torch.log2(z_likelihood), dim=(1,2,3)).\
              reshape(b, -1) / (h * w)
    local_d = IWAELoss(d_loss)
    local_yz = IWAELoss(yz_loss)
    local_z = IWAELoss(z_loss)
    loss_total = local_d + local_yz + local_z

    return loss_total
\end{lstlisting}
\end{small}
\subsection{Detailed Experimental Results}
\label{ap:detailsexp}

In this section we present more detailed experimental results in Tab.~\ref{tab:detailgg18} and Tab.~\ref{tab:cheng20results}. Note that without \textit{direct-y} trick, the IWAE for \citet{cheng2020learned} totally fails and we can not produce a valid BD metric from it.

\begin{table}[ht]
\centering
\caption{Detailed results based on \citep{balle2018variational}.}
\vspace{2mm}
\label{tab:detailgg18}
\begin{tabular}{lccccc}
\toprule
                      & $\lambda$  & bpp & MSE & PSNR (db) & MS-SSIM \\ \midrule
Baseline \citep{balle2018variational}
                & 0.0016 & 0.1205 & 138.4 & 27.23 & 0.9111 \\
                & 0.0032 & 0.1990 & 91.52 & 28.95 & 0.9384 \\
                & 0.0075 & 0.3492 & 52.68 & 31.28 & 0.9624 \\
                & 0.015 & 0.5270 & 32.78 & 33.28 & 0.9766 \\
                & 0.03 & 0.7626 & 19.90 & 35.37 & 0.9847 \\
                & 0.045 & 0.9249 & 15.69 & 36.39 & 0.9883 \\
                & 0.08 & 1.211 & 10.04 & 38.27 & 0.9919 \\\midrule
IWAE \citep{burda2016importance}      
                & 0.0016 & 0.2559 & 144.7 & 27.12 & 0.9134 \\
                & 0.0032 & 0.3478 & 90.65 & 29.03 & 0.9389 \\
                & 0.0075 & 0.5931 & 51.40 & 31.38 & 0.9642 \\
                & 0.015 & 0.7902 & 31.73 & 33.40 & 0.9765 \\
                & 0.03 & 1.135 & 19.41 & 35.47 & 0.9850 \\
                & 0.045 & 1.886 & 14.70 & 36.65 & 0.9885 \\
                & 0.08 & 1.753 & 9.898 & 38.32 & 0.9919 \\\midrule
MS-NIC-MIX      & 0.0016 & 0.1132 & 146.4 & 27.08 & 0.9121 \\
                & 0.0032 & 0.1967 & 88.44 & 29.15 & 0.9409 \\
                & 0.0075 & 0.3496 & 51.27 & 31.39 & 0.9632 \\
                & 0.015 & 0.5260 & 31.52 & 33.43 & 0.9773 \\
                & 0.03 & 0.7591 & 19.33 & 35.49 & 0.9851 \\
                & 0.045 & 0.9248 & 14.43 & 36.72 & 0.9885 \\
                & 0.08 & 1.201 & 9.694 & 38.40 & 0.9919 \\\midrule
MS-NIC-DMS      & 0.0016 & 0.1173 & 135.1 & 27.38 & 0.9145 \\
                & 0.0032 & 0.1967 &	86.07 & 29.26 & 0.9413 \\
                & 0.0075 & 0.3495 & 49.98 & 31.51 & 0.9647 \\
                & 0.015 & 0.5250 & 31.36 & 33.46 & 0.9771 \\
                & 0.03 & 0.7546 & 19.81 & 35.37 & 0.9846 \\
                & 0.045 & 0.9220 & 14.74 & 36.61 & 0.9883 \\
                & 0.08 & 1.196 & 9.637 & 38.43 & 0.9920 \\\bottomrule
\end{tabular}
\end{table}

\begin{table}[ht]
\centering
\caption{Detailed results based on \citep{cheng2020learned}.}
\vspace{2mm}
\label{tab:detailcheng20}
\begin{tabular}{lccccc}
\toprule
                      & $\lambda$  & bpp & MSE & PSNR (db) & MS-SSIM \\ \midrule
Baseline \citep{cheng2020learned}
                & 0.0016 & 0.1205 & 138.4 & 27.23 & 0.9111 \\
                & 0.0032 & 0.1990 & 91.52 & 28.95 & 0.9384 \\
                & 0.0075 & 0.3492 & 52.68 & 31.28 & 0.9624 \\
                & 0.015 & 0.5270 & 32.78 & 33.28 & 0.9766 \\
                & 0.03 & 0.6424 & 19.48 & 35.54 & 0.9855 \\
                & 0.045 & 0.7846 & 15.48 & 36.53 & 0.9885 \\
                & 0.08 & 1.026 & 11.41 & 37.87 & 0.9916 \\\midrule
IWAE \citep{burda2016importance}      
                & 0.0016 & 3.226 & 109.1 & 28.32 & 0.9182 \\
                & 0.0032 & 3.407 & 78.07 & 29.74 & 0.9414 \\
                & 0.0075 & 3.555 & 47.19 & 31.84 & 0.9652 \\
                & 0.015 & 3.445 & 31.84 & 33.56 & 0.9779 \\
                & 0.03 & 3.534 & 23.66 & 34.92 & 0.9849 \\
                & 0.045 & 3.545 & 20.63 & 35.59 & 0.9878 \\
                & 0.08 & 3.157 & 16.40 & 36.62 & 0.9908 \\\midrule
MS-NIC-MIX      & 0.0016 & 0.1068 & 109.7 & 28.30 & 0.9171 \\
                & 0.0032 & 0.1636 & 78.07 & 29.71 & 0.9404 \\
                & 0.0075 & 0.2861 & 47.29 & 31.85 & 0.9651 \\
                & 0.015 & 0.4309 & 31.88 & 33.55 & 0.9777 \\
                & 0.03 & 0.6586 & 19.11 & 35.60 & 0.9853 \\
                & 0.045 & 0.8007 & 14.65 & 36.76 & 0.9889 \\
                & 0.08 & 1.034 & 10.93 & 38.00 & 0.9916 \\\midrule
MS-NIC-DMS      & 0.0016 & 0.1043 & 109.8 & 28.30 & 0.9163 \\
                & 0.0032 & 0.1644 & 77.44 & 29.74 & 0.9412 \\
                & 0.0075 & 0.2849 & 47.30 & 31.85 & 0.9656 \\
                & 0.015 & 0.4306 & 32.22 & 33.51 & 0.9775 \\
                & 0.03 & 0.6432 & 18.94 & 35.65 & 0.9856 \\
                & 0.045 & 0.7926 & 14.84 & 36.67 & 0.9886 \\
                & 0.08 & 1.039 & 10.48 & 38.18 & 0.9918\\\bottomrule
\end{tabular}
\end{table}

\subsection{Distribution of Latent Variance}
\label{app:dlv}
\begin{figure}[htb]
\begin{center}
    \includegraphics[width=\linewidth]{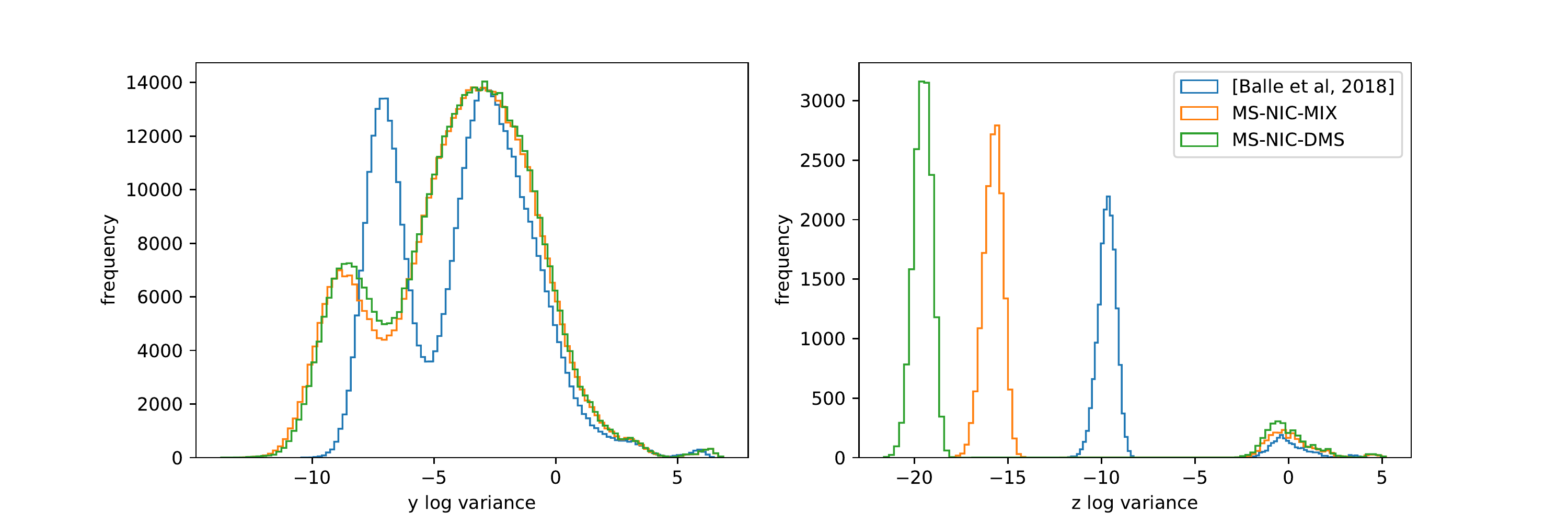}
    \label{fig:hst}
    \caption{The histogram of log space per-dimension latent variance across Kodak test images. The model is trained with $\lambda = 0.015$.}
\end{center}
\end{figure}

We show the histogram of latent variance in log space in Fig.~\ref{fig:hst}. From the histogram we can observe that for latent $\bm{y}$, the variance distribution of two MS-NIC approaches is similar and single-sample approach is quite different. MS-NIC has more latent dimensions that have high variance (the right mode), and less with low variance (the left mode). Moreover, the low variance mode of MS-NIC has less variance than single-sample approach, which indicates that MS-NIC does a better job in separating active and inactive latent dimensions. Similarly, the low variance mode of $\bmt{z}$ in MS-NIC approaches is lower than single-sample approach.
\begin{table}[ht]
\centering
\caption{The average of per-dimension latent variance and Cov across Kodak test images. The model is trained with $\lambda = 0.015$.}
\vspace{2mm}
\label{tab:nord}
\begin{tabular}{lcccc}
\toprule
                     & \multicolumn{2}{c}{Var(\#)} & \multicolumn{2}{c}{Cov(\#)} \\ \cmidrule(l){2-3} \cmidrule(l){4-5}
                       Method & $\bm{y}$ & $\bm{z}$ & $\bm{y}$ & $\bm{z}$ \\ \midrule
\textit{Single-sample} & &  \\
\citet{balle2018variational} & 1.512 & 0.3356 & 20.36 & 14.48 \\ \midrule
\textit{Multiple-sample} &&&& \\
MS-NIC-MIX & 1.909 & 0.7705 & 114.4 & 7.234  \\
MS-NIC-DMS & 1.908 & 0.7522 & 93.67 & 7.024 \\\bottomrule
\end{tabular}
\end{table}

\subsection{Tighter ELBO for Inference Time}
\label{app:teit}
\subsubsection{Inference time ELBO and Softmin Coding \citep{theis2021importance}}
The inference time tighter ELBO is another under-explored issue. In fact, the training time tighter ELBO and inference time ELBO is independent. We can train a model with tighter ELBO, infer with single sample ELBO. Or we can also conduct multiple sample infer on a model trained with single sample. The general idea is:
\begin{itemize}
    \item The training time tighter ELBO benefits the performance in terms of avoiding posterior collapse. As we state and empirically verify in Sec. 5.3. We adopt deterministic rounding during inference time, and there is no direct connection between the training time tighter ELBO and inference time R-D cost. However, we indeed end up with a richer latent space (Sec. 5.3), which means more active latent dimensions and less bitrate waste.
    \item The inference time tighter ELBO sounds really alluring for compression community. However, there remains two pending issue to be resolved prior to the application of the inference time tighter ELBO: 1) How this inference time multiple-sample ELBO is related to R-D cost remains under-explored. In other words, whether the entropy coding itself can achieve the R-D cost defined by multiple-sample ELBO is a question. 2) The inference time multiple-sample ELBO only makes sense with stochastic encoder (you can not importance weight the same deterministic ELBO), whose impact on lossy compression remains dubious. 
\end{itemize}
For the first pending issue, the softmin coding \citep{theis2021importance} is proposed to achieve multiple sample ELBO based on Universal Quantization (UQ) \citep{agustsson2020universally}. However, it is not a general method and is tied to UQ. Moreover, as we stated in Sec 4.3, its computational cost is forbiddingly high and its improvement is marginal. But those are not the real problem of softmin coding. Instead, the real problem is the second pending issue: stochastic lossy encoder. The softmin coding relies on UQ, and UQ relies on stochastic lossy encoder. And the stochastic lossy encoder is exactly the second issue that we want to discuss.
\subsubsection{Stochastic Lossy Encoder and Universal Quantization}
It is known to lossless compression community that stochastic lossy encoder benefits compression performance \citep{ryder2022split} with the aid of bits-back coding \citep{townsend2018practical}. While the bits-back coding is not applicable to lossy compression. For lossy compression, currently we know that the stochastic encoder degrades R-D performance especially when distortion is measured in MSE \citep{theis2021advantages}. In the original UQ paper, the performance decay of vanilla UQ over deterministic rounding is obvious ($\approx 1$db). When we writing this paper, we also find the performance decay of UQ is quite high. As shown in Tab.~\ref{tab:uqrd}, the R-D cost of UQ is significantly higher than deterministic rounding. This negative result makes softmin coding less promising than it seems as it only obtains a marginal gain over UQ.
\begin{table}[htb]
\begin{center}
\begin{tabular}{@{}lllll@{}}
\toprule
 & y bpp & z bpp & MSE & RD Cost \\ \midrule
Deterministic Rounding & 0.3347 & 0.01418 & 26.86 & 0.7552 \\
Universal Quantization & 0.5379 & 0.01431 & 23.94 & 0.9080 \\ \bottomrule
\end{tabular}
\end{center}
\caption{The R-D performance of UQ vs deterministic rounding on the first image of Kodak dataset.}
\label{tab:uqrd}
\end{table}
In our humble opinion, this performance decay of UQ is partially brought by stochastic encoder itself. For lossless compression, the deterministic encoder and stochastic encoder are just two types of bit allocation preference:
\begin{itemize}
    \item The deterministic encoder allocate less bitrate to $\log p(y)$, more to $\log p(x|y)$ and $0$ to $\log q(y|x)$.
    \item The stochastic encoder allocate more bitrate to $\log p(y)$, less to $\log p(x|y)$ and minus bitrate to $\log q(y|x)$
\end{itemize}
Therefore, for lossless compression, it is reasonable that the bitrate increase to $\log p(y)$ and $\log p(x|y)$ can be offset by bits-back coding bitrate $\log q(y|x)$. While for lossy compression, there is no way to bits-back $\log q(y|x)$ (as we can not reconstruct $q(y|x)$ without $x$). If the bitrate increase in $\log p(y)$ and $\log p(x|y)$, the R-D cost just increases for lossy compression. Prior to other entropy coding bitrate that is able to achieve R-D cost equals to minus ELBO with $E_q[\log q]\neq 0$ becomes mature (such as relative entropy coding \citep{flamich2020compressing}), we have no way to implement a stochastic lossy encoder with reasonable R-D performance. By now, we have no good way to achieve tighter ELBO during inference time.
\subsubsection{Training-Testing Distribution Mismatch and Universal Quantization}
\begin{figure}[htb]
\begin{center}
    \includegraphics[width=0.8\linewidth]{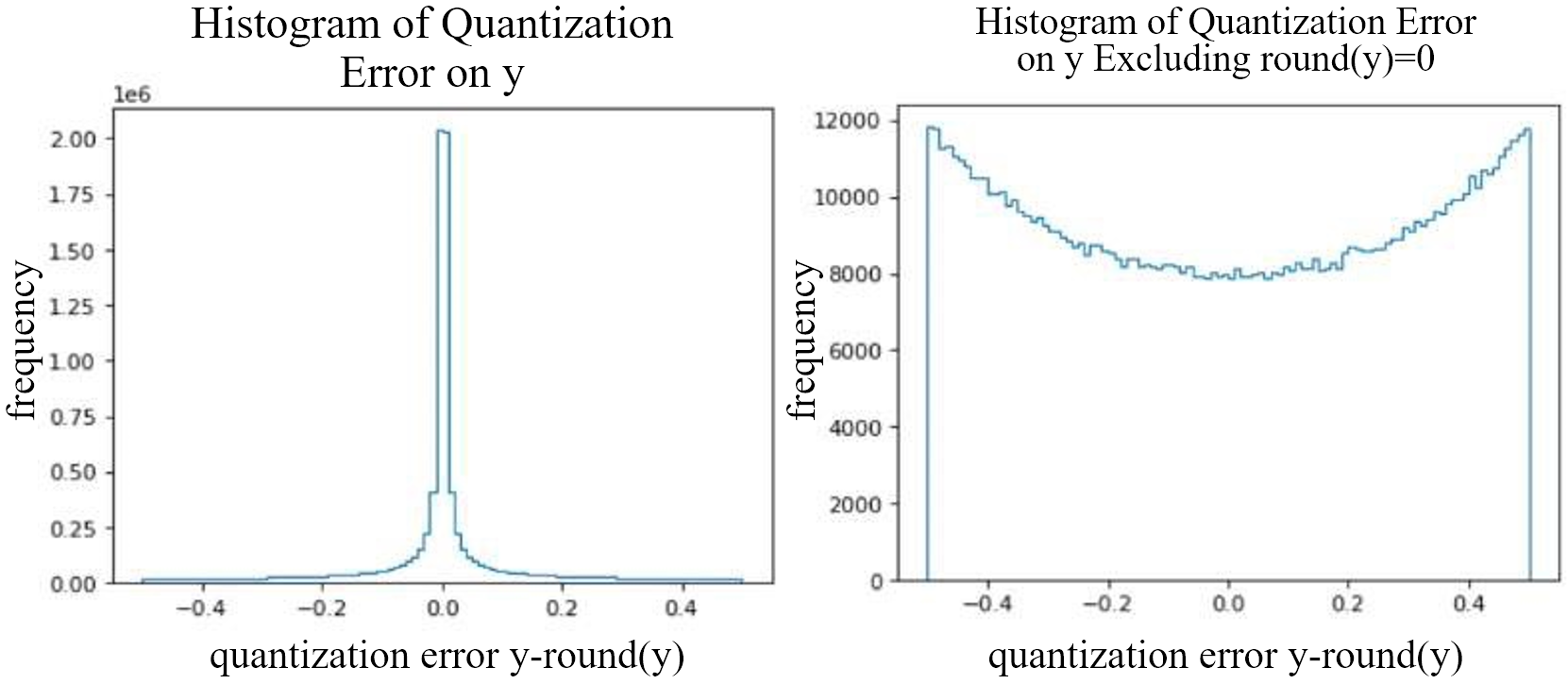}
    \label{fig:hstqe}
    \caption{The histogram of $\bm{y}-\bar{\bm{y}}$ of first image of Kodak dataset.}
\end{center}
\end{figure}

Moreover, whether the quantization error is uniform distribution remains a question. And we think that is another reason why UQ does not work well. In fact, the real distribution of quantization noise is pretty much a highly concentrated distribution around $0$ (See Fig.~\ref{fig:hstqe}). And it is quite far away from uniform distribution, which violates the assumption of \citet{balle2018variational}. We also find that this concentrated distribution is caused by that most of latent dimension is quite close to $0$. The evidence is, if we remove the latent dimension $y^i\in [-0.5,0.5]$, then the quantization noise looks similar to a uniform distribution (See Fig.~\ref{fig:hstqe}). So, if we apply direct rounding, they are kept as $0$ and the latent is sparse. However, adding uniform noise to it loses this sparsity, which result in bitrate increase. And from Tab.~\ref{tab:uqrd}, we can wee that the UQ reduce MSE while increase the bitrate. From total R-D cost perspective, the deterministic rounding outperforms UQ. As a matter of fact, the assumption of UQ that resolving training-testing distribution mismatch improves R-D performance does not hold well. To wrap up, we find that there is some pending issues to be resolved prior to the practical solution of tighter ELBO for inference time.

\subsection{The Effect on Training Time} The MS-NIC-MIX and MS-NIC-DMS is more time efficient than simply increase batchsize. For MS-NIC-MIX with $k$ samples, the $y$ encoder $q(y|x)$ is inferred with only 1 sample, and the $z$ encoder, decoder and entropy model $q(z|x),p(y|z),p(z)$ is inferred with only 1 sample. And only the $y$ decoder $p(x|y)$ is inferred $k$ times. This sample efficiency makes the training time grows slowly with $k$. In our experiment, the MS-NIC-MIX with 8 samples only increases the training time by $\times 1.5$, the MS-NIC-MIX with $16$ samples only increases the training time by $\times 3$. The MS-NIC-DMS is slightly slower, as the $z$ entropy model and decoder $p(z),p(y|z)$ also requires $k$ times inference. However, it is still much more efficient than batchsize $\times k$ as all the encoders $q(y|x),q(z|x)$ requires only 1 inference. In fact, sampling from posterior is much cheaper than inferring the posterior parameters. Similar spirit has also been adopted in improving the efficiency of sampling from Gumbel-Softmax relaxed posterior 
\citep{paulus2020rao}.

The trade-off between batchsize and sample number is a more subtle issue. As stated in \citet{rainforth2018tighter}, the gradient SNR of encoder (inference model) scales with $\Theta(M/K)$, and the gradient SNR of decoder (generative model) scales with $\Theta(M/K)$, where M is the batchsize and K is the sample size. Another assumption required prior to further discussion is that the suboptimality of VAE mainly comes from inference model \citep{cremer2018inference}, which means that the encoder is harder to train than the decoder. This means that an infinitely large $K$ ruins the convergence of encoder, and solemnly increasing sample number frustrates training. In practice the overall performance is determined by both inference suboptimality and ELBO-likelihood gap. In a word, we believe there is no general answer for all problem. But a reasonable balance between sample size and batchsize is the golden rule to maximize performance (as $T(M)$ and $T(K)$ grow linearly with batchsize/sample size). And the obvious case is that neither setting batchsize to $M=1$ and give all resources to $K$, nor setting sample size $K=1$ and give all resources to $M$ is optimal.

\subsection{More Limitation and Discussion}
\label{app:disc}

The cause of negative results on MS-SSIM of \citet{cheng2020learned} is more complicated. One possible explanation is that the gradient property of \citet{cheng2020learned} is not as good as \citet{balle2018variational}. As a reference, the training of \citep{burda2016importance} totally fails on \citet{cheng2020learned} and produces garbage R-D results (See Tab.~\ref{tab:detailcheng20}). This bad gradient property might account for the bad results of MS-SSIM on \citet{cheng2020learned}, as the gradient of IWAE and MS-NIC is certainly trickier than the gradient of single sample approaches.

As evidence, when we are studying the stability of the network in \citet{cheng2020learned}, we find that without limitation of entropy model (imagine setting $\lambda$ to $\infty$) and quantization, \citet{cheng2020learned} produces PSNR of $43.27$db, while \citet{balle2018variational} produces PSNR of $48.54$db. This means that \citet{cheng2020learned} is not as good as \citet{balle2018variational} as an auto-encoder. Moreover, when we finetune these pre-trained model into a lossy compression model, \citet{cheng2020learned} produces $nan$ results while \citet{balle2018variational} converges. This result indicates that the backbone of \citet{cheng2020learned}'s gradient is probably more difficult to deal with than \citet{balle2018variational}.

\subsection{Broader Impact}

Improving the R-D performance of NIC methods is valuable itself. It is beneficial to reducing the carbon emission by reducing the resources required to transfer and store images. And NIC has potential of saving network channel bandwidth and disk storage over traditional codecs. Moreover, for traditional codecs, usually dedicated hardware accelerators are required for efficient decoding. This codec-hardware bondage hinders the wide adaptation of new codecs. Despite the sub-optimal R-D performance of old codecs such as JPEG, H264, they are still prevalent due to broad hardware support. While modern codecs such as H266 \citep{bross2021developments} can not be widely adopted due to limited hardware decoder deployment. However, for NIC, the general purpose neural processors are able to fit to all codecs. Thus the neural decoders have better hardware flexibility, and the cost to update neural decoder only involves software, which encourages the adoption of newer methods with better R-D performance.

\end{document}